\documentclass[10pt]{article} 
\usepackage[preprint]{rlc}

\usepackage{algcompatible}
\usepackage{algorithm}
\usepackage{makecell}
\usepackage{xcolor}
\usepackage{adjustbox}
\usepackage{graphicx}
\usepackage{subcaption}
\usepackage{svg}
\usepackage{amsmath}
\usepackage{amssymb}
\usepackage{amsthm}
\usepackage{siunitx}
\usepackage{float}
\usepackage{wrapfig}
\usepackage{tikz}
\usetikzlibrary{calc}
\usepackage{hyperref}
\usepackage{url}

\usepackage{xcolor}
\definecolor{plotred}{HTML}{CA3720}
\definecolor{plotblue}{HTML}{0086DA}
\definecolor{plotgreen}{HTML}{2BAF2A}
\definecolor{plotpurple}{HTML}{9923DC}
\definecolor{plotnblue}{HTML}{34495E}

\newcommand{\plotfontvsmall}{\fontsize{7}{9}\selectfont}
\newcommand{\plotfontksmall}{\fontsize{8}{10}\selectfont}
\newcommand{\plotfontsmall}{\fontsize{9}{11}\selectfont}
\newcommand{\plotfont}{\fontsize{10}{12}\selectfont}
\newcommand{\defeq}{\mathrel{%
    \stackrel{\raisebox{-0.9ex}{$\cdot$}}{=}%
}}

\DeclareMathOperator{\btheta}{\boldsymbol{\theta}}
\DeclareMathOperator{\bphi}{\boldsymbol{\phi}}
\DeclareMathOperator{\bpsi}{\boldsymbol{\psi}}

\DeclareMathOperator*{\argmin}{arg\,min}

\title{Harnessing Discrete Representations for\\
Continual Reinforcement Learning}

\author{Edan Meyer \\
    ejmeyer@ualberta.ca
    \And
    Adam White* \\
    amw8@ualberta.ca
    \And
    Marlos C. Machado* \\
    machado@ualberta.ca
    \and
    \\
    \begin{minipage}{\textwidth}
    \centering
    Alberta Machine Intelligence Institute (Amii) \\
    Department of Computing Science, University of Alberta \\
    *Canada CIFAR AI Chair
    \end{minipage}
}

\begin{document}

\maketitle

\begin{abstract}
Reinforcement learning (RL) agents make decisions using nothing but observations from the environment, and consequently, rely heavily on the representations of those observations. Though some recent breakthroughs have used vector-based categorical representations of observations, often referred to as discrete representations, there is little work explicitly assessing the significance of such a choice. In this work, we provide an empirical investigation of the advantages of discrete representations in the context of world-model learning, model-free RL, and ultimately continual RL problems, where we find discrete representations to have the greatest impact. We find that, when compared to traditional continuous representations, world models learned over discrete representations accurately model a larger portion of the state space with less capacity, and that agents trained with discrete representations learn better policies with less data. In the context of continual RL, these benefits translate into faster adapting agents. Additionally, our analysis suggests that it is the binary and sparse nature, rather than the ``discreteness'' of discrete representations that leads to these improvements.\footnote{Code for the implementation and analysis is accessible at\\\url{https://github.com/ejmejm/discrete-representations-for-continual-rl}.}

\end{abstract}

\section{Introduction} \label{sec: introduction}

This work is motivated by the quest to design autonomous agents that can learn to achieve goals in their environments solely from their stream of experience. The field of reinforcement learning (RL) models this problem as an agent that takes actions based on observations of the environment in order to maximize a scalar reward. Given that observations are the agent's sole input when choosing an action (unless one counts the history of reward-influenced policy updates), the representation of observations plays an indisputably important role in RL.


In this work, we examine the understudied yet highly effective technique of representing observations as vectors of categorical values, referred to in the literature as discrete representations \citep{vqvae_2018, dreamerv2_2022, disentangled_disc_reprs_2023} --- a method that stands in contrast to the conventional deep learning paradigm that operates on learning continuous representations. Despite the numerous uses of learned, discrete representations \citep[e.g.,][]{smaller_wm_2021, dreamerv3_2023, iris_2023}, the mechanisms by which they improve performance are not well understood. To our knowledge, the only direct comparison to continuous representations in RL comes from a single result from \citet{dreamerv2_2022} in a subfigure in their paper. In this work, we dive deeper into the subject and investigate the effects of discrete representations in RL.

The successes of discrete representations in RL date back to at least as early as tile coding methods, which map observations to multiple one-hot vectors via a hand-engineered representation function \cite[p. 217-222]{rl_book_2018}. Tile coding was most popular prior to the proliferation of deep neural networks as a way to construct representations that generalize well.
Continuous alternatives existed --- notably, radial basis functions (RBFs) could be viewed as a generalization of tile coding that produce values in the interval $[0, 1]$. But despite the superior representational capacity of RBFs, they have tended to underperform in complex environments with high-dimensional observations \citep{an_1991, lane_1992}.

A similar comparison can be seen between the work of \citet{dqn_nature_2015} and \citet{shallow_atari_2016}. \citeauthor{dqn_nature_2015} train a deep neural network (DNN) to play Atari games, relying on the neural network to learn its own useful representation, or features, from pixels. In contrast, \citeauthor{shallow_atari_2016} construct a function for producing binary feature vectors that represent the presence of various patterns of pixels, invariant to position and translation. From this representation, a linear function approximator is able to perform as well as a DNN trained from pixels.

Recent approaches to producing discrete representations in the area of supervised learning have moved away from hand-engineering representations, and towards learning representations. \Citet{vqvae_2018}, for example, propose the vector quantized variational autoencoder (VQ-VAE), a self-supervised method for learning discrete representations. VQ-VAEs perform comparably to their continuous counterparts, variational autoencoders \citep{vae_2014}, while representing observations at a fraction of the size. When applied to DeepMind Lab \citep{deepmind_lab_2016}, VQ-VAEs are able to learn representations that capture the salient features of observations, like the placement and structure of walls, with as little as 27 bits \citep{vqvae_2018}.

Similar representation learning techniques have also been successfully applied in the domain of RL. \citet{dreamerv2_2022} train an agent on Atari games \citep{atari_2013, revisiting_ale_2018}, testing both discrete and continuous representations. They find that agents learning from discrete representations achieve a higher average reward, and carry on the technique to a follow-up work \citep{dreamerv3_2023} where they find success in a wider variety of domains, including the Proprio Control Suite \citep{dm_control_suite_2018}, Crafter \citep{crafter_2022}, and Minecraft \citep{malmo_2016}. Works like those from \citet{smaller_wm_2021} and \citet{iris_2023} further build on these successes, using discrete representations to learn world models and policies. Work from \citet{fta_repr_2023} finds that representations that are more successful in transfer learning are often sparse and orthogonal, suggesting that these properties may underpin such successes of discrete representations.

The goal of this work is to better understand how discrete representations help RL agents. We use vanilla autoencoders \citep{autoencoder_1987} to learn dense, continuous representations, fuzzy tiling activation (FTA) autoencoders \citep{fta_2021} to learn sparse, continuous representations, and vector quantized-variational autoencoders (VQ-VAEs) to learn fully discrete, binary representations. Inspired by the success of the Dreamer architecture \citep{dreamerv2_2022, dreamerv3_2023}, we first examine how these different representations help in two distinct parts of a model-based agent: world-model learning and (model-free) policy learning. Observing that discrete and sparse representations specifically help when an agent's resources are limited with respect to the environment, we turn to the continual RL setting, where an agent must continually adapt in response to its constrained resources \citep{cl_as_constrained_rl_2023}. We particularly emphasize the benefits of discrete and sparse representations in continual RL, as the largest and most complex environments are impossible to perfectly model and require continual adaptation to achieve the best possible performance \citep{tracking_vs_converging_2007, alberta_plan_2022}.

The primary contributions of our work include:
\vspace{-0.5em}
\begin{itemize}
    \item Showing that discrete representations can help learn better models and policies with less resources (modeling capacity and data).
    \item Demonstrating that the successes of discrete representations are likely attributable to the choice of one-hot encoding rather than the ``discreteness'' of the representations themselves. 
    \item Identifying and demonstrating that discrete and sparse representations can help continual RL agents adapt faster.
\end{itemize}

\section{Experimental Setup} \label{sec: background}

\looseness=-1
This work primarily focuses on how to train agents to achieve some goal in their environment by learning to select actions, $A_t \in \mathcal{A}$. This problem is formulated as learning to select actions from states $S_{t} \in \mathcal{S}$, that best maximize a given reward signal, $R_{t+1} \in \mathbb{R}$. We are specifically concerned with how to learn the parameters, $\btheta$, of a policy,  $\pi_{\btheta}(A_t | S_t)$, that maps from states to a distribution over actions. The goal is to maximize the discounted return from the current state, which is given by $G_t\defeq\sum_{k=0}^{T}{\gamma^k R_{t+k+1}}$, where $T$ is the terminal time step, and $\gamma \in [0, 1]$ is the discount factor. In this following parts of this section, we discuss the algorithms we use to achieve this goal, and the environments we use in our experiments.

\subsection{Algorithms}

\looseness=-1
We use proximal policy optimization (PPO) \citep{ppo_2017} to learn policies, which collects transitions through environment interactions, and then applies multiple epochs of stochastic gradient descent to weights that directly parameterize the policy. The sample efficiency of model-free RL algorithms like PPO can sometimes be further improved with the additional use of a world model \citep{direct_mbrl_comparison_1997, linear_dyna_2008, q_learning_efficient_2018, mbpo_2019}. In our work, we independently study two components that are often part of model-based RL methods---world-model learning and (model-free) policy learning---for a fine-grained view of how the different types of representations affect complex RL agents.

Both policy and world model architectures are split into two components in our work: a representation network (or encoder) that extracts a representation, and a higher-level network that learns a policy or world model atop the learned representations. This decoupling allows us to swap out the encoder (both architecture and objective), while keeping the higher-level model unchanged. With the exception of an end-to-end baseline, each of the encoders we use are trained with an observation reconstruction objective as part of a larger autoencoder model \citep{autoencoder_1987}. The autoencoder architecture compresses an observation into a bottleneck state before attempting to reconstruct it, forcing it to learn a representation that captures salient aspects of the observation. Each of the three types of learned representations in our work are produced by different autoencoder variants. We also evaluate the standard approach of end-to-end learning, where the representations are learned as a byproduct of the optimization process.

Dense, continuous representations are produced by a vanilla autoencoder.\footnote{We also tested variational autoencoders \citep{vae_2014} in early model learning experiments, but were unable to find hyperparameters to make the method competitive.} Sparse, continuous representations also use a vanilla autoencoder, but the bottleneck layer outputs are passed through a Fuzzy Tiling Activation (FTA) \citep{fta_2021}. FTA produces sparse outputs by converting scalars to ``fuzzy'' one-hot vectors. The FTA representations provide a strong baseline \citep{erfan_thesis, fta_repr_2023} that acts as a bridge between dense, continuous representations and discrete representations. Discrete representations are produced by a vector quantized-variational autoencoder (VQ-VAE) \citep{vqvae_2018}, which quantizes the multiple outputs of the encoder to produce a vector of discrete values, also referred to as the \textit{codebook}. The discrete representations we refer to in our work comprise multiple one-hot vectors, each representing a single, discrete value from the codebook. The details of these autoencoders are explained in more depth in Appendix~\ref{sec:ae_preliminaries}.

\subsection{Environments}

\begin{wrapfigure}{r}{0.53\textwidth}  
    \begin{subfigure}[b]{0.15\textwidth}
        \includegraphics[width=\textwidth]{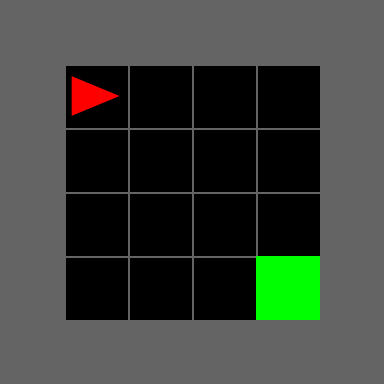}
        \caption{Empty}
        \label{fig:empty_env}
    \end{subfigure}
    \hfill
    \begin{subfigure}[b]{0.15\textwidth}
        \includegraphics[width=\textwidth]{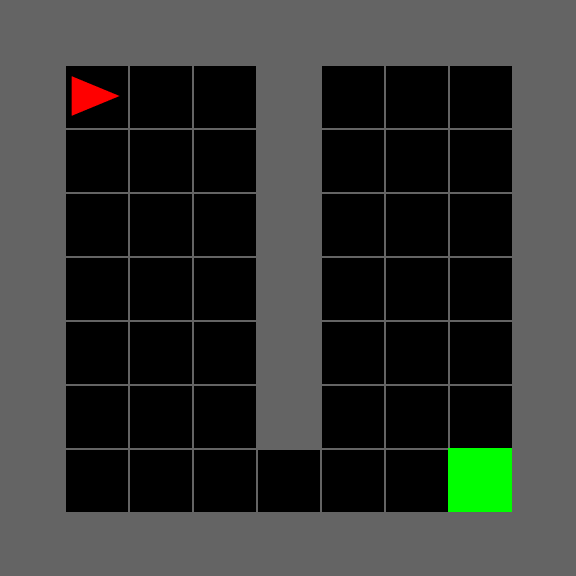}
        \caption{Crossing}
        \label{fig:crossing_env}
    \end{subfigure}
    \hfill
    \begin{subfigure}[b]{0.15\textwidth}
        \includegraphics[width=\textwidth]{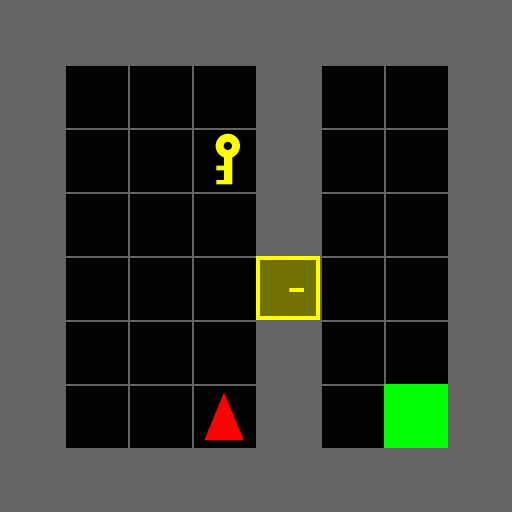}
        \caption{Door Key}
        \label{fig:door_key_env}
    \end{subfigure}
    \caption{Minigrid environments used in our experiments. We refer to these as the (a) \textit{empty}, (b) \textit{crossing}, and (c) \textit{door key} environments. The agent receives lower-resolution RGB arrays representing pixels as observations.}
    \label{fig:envs}
\end{wrapfigure}

Throughout this work, we use the \textit{empty}, \textit{crossing}, and \textit{door key} Minigrid environments \citep{minigrid_2023}, as displayed in Figure \ref{fig:envs}. In each environment, the agent receives pixel observations, and controls a red arrow that navigates through the map with \texttt{left}, \texttt{right}, and \texttt{forward} actions. The agent in the \textit{door key} environment additionally has access to \texttt{pickup} and \texttt{use} actions to pickup the key and open the door. The \textit{crossing} and \textit{door key} environments are stochastic, with each action having a 10\% chance to enact a random, different action. The stochasticity increases the difficulty of learning a world model by increasing the effective number of transitions possible in the environments. The increase in difficulty widens the performance gap between different methods, which makes the results easier to interpret.

The environments are episodic, terminating when the the agent reaches the green square, or when the episode reaches a maximum length. The former yields a reward $R_t \in [0.1, 1]$ depending on the length of the episode (shorter episodes yield higher rewards), and the latter yields no reward. The reward is calculated with the standard Minigrid formula, $1 - 0.9 \frac{t}{T}$, where $t$ is the current step and $T$ is the maximum episode length (dependent on the experiment). Contrary to the standard Minigrid environments, the layouts are fixed throughout all episodes. Further environment details are displayed in Table~\ref{table:env_stats} in Appendix~\ref{sec:rl_hyperparams}.

\section{World-Model Learning with Discrete Representations} \label{sec:model_learning}

We begin our experiments by examining the benefits of using discrete representations in world model learning. We specifically focus on the case of sample models, where the model is trained to produce outcomes with probability equal to that of outcomes in the environment.

\subsection{Learning World Models} \label{sec:model_training}

We train autoencoders and world models on a static dataset, $\mathcal{D}$, of one million transition tuples, $(s, a, s')$, collected with random walks. In each episode, the environment terminates when the agent reaches the green square or after 10,000 steps.  Training occurs in two phases: first the autoencoder is trained, and then a transition model is trained over the fixed representations.

Observations are 3-dimensional RGB arrays, so we use convolutional and deconvolutional neural networks \citep{cnn_1989} for the encoder and decoder architectures. The encoder architecture is similar to the IMPALA network \citep{impala_2018}, but the size of the bottleneck layer is chosen with a hyperparameter sweep. Architectural details are given in Section \ref{sec:ae_architecture}. All of the autoencoders are trained with a mean squared error reconstruction loss, and the VQ-VAE with additional loss terms as detailed in Section \ref{sec:ae_preliminaries}. Training for both autoencoders and world models use the Adam optimizer \citep{adam_2014} with hyperparameter values of $\beta_1 = 0.9$, $\beta_2 = 0.999$, and a step size of $\num{2e-4}$.  Training continues for a fixed number of epochs, until near-convergence, at which point the model weights are frozen and world model learning begins.

World models learned over latent representations take a latent state, $\mathbf{z}$, and an action, $a$, as input to predict the next latent state, $\mathbf{\hat{z}'} = w_{\mathbf{\psi}}(\mathbf{z}, a)$, with an MLP, $w_{\bpsi}$. World models learned over continuous representations, or \textit{continuous world models}, consist of three layers of 64 hidden units (32 in the \textit{crossing} environment), and rectified linear units (ReLUs) \citep{relu_2018} for activations. In \textit{discrete world models}, the MLP is preceded by an embedding layer that converts discrete values into a continuous, 64-dimensional vectors. The loss for both world models is given by the difference between the predicted next latent state and the ground-truth next latent state. The continuous world model outputs a continuous vector and uses the squared error loss. The discrete model outputs multiple vectors of categorical logits and uses a categorical cross-entropy loss over each.\footnote{We also experimented with a squared error loss for the discrete world model and found it made little difference in the final world model accuracy.} All world models are trained with 4 steps of hallucinated replay as described by \citet{halucinated_replay_2017}. Hallucinated replay entails feeding outputs of the model back in as new inputs, and training over multiple ``hallucinated'' steps to increase the accuracy of the world model. Figures \ref{fig:cont_model_viz} and \ref{fig:disc_model_viz} in Appendix~\ref{sec:appendix_world_model} depict the training process for continuous and discrete world models, and include a visualization of hallucinated replay.

Our aim is to train sample models---models that emulate the environment by producing outcomes with frequency equivalent to that of the real environment. This is more difficult in stochastic environments because our current training procedure would result in expectations models, where predictions are weighted averages over possible outcomes. To instead learn sample models, we augment our models using the method proposed by \citet{stochastic_muzero_2022}. This approach learns a distribution over potential outcomes, and samples from it when using the world model. We provide a more detailed explanation and relevant hyperparameters in Appendix \ref{sec:stoch_world_models}.

\subsection{Experiments}

\looseness=-1
The goal of this first set of experiments is to measure \textbf{how the representation of the latent space affects the ability to learn an accurate world model}. Unfortunately, this is not as simple as comparing a predicted latent state to the ground-truth latent state, as multiple outcomes may be possible for any given state-action pair. To account for this, we look at distributions over many transitions instead of the outcomes of single transitions. Specifically, we choose a behavior policy and measure the difference between the state distribution it induces in the real environment and in a learned model of the environment. Accurate world models should produce state distributions similar to that of the real environment, and inaccurate models should produce state distributions that differ. Figure \ref{fig:crossing_state_distrib} in Appendix~\ref{sec:appendix_world_model} contains a visualization that helps build an intuition of how state distributions may differ, which we will discuss in more detail later.

\looseness=-1
Emulating how world models are often used to simulate multiple different policies, we choose different behavior policies in each environment. We use a random policy for the \textit{empty} environment, a policy that explores the right half of the grid in the \textit{crossing} environment, and a policy that navigates directly to the goal in the \textit{door key} environment. We enact the policies in the real environments and learned world models for 10,000 episodes each. Episodes are cut off early, or are frozen at the terminal state to reach exact 30 steps of interaction. We then compare the KL divergence between ground-truth and induced state distributions at each step of the rollouts. A lower KL divergence is better, indicating that a model predicts outcomes more similar to the real environment.

\looseness=-1
We include two baselines in our comparisons that are free of auxiliary autoencoder objectives: the uniform baseline and the end-to-end baseline. The uniform baseline predicts a uniform distribution over all states and is strong when the agent's target policy leads it to spread out, like in a random walk. The end-to-end baseline shares an architecture equivalent to the vanilla autoencoder, but is trained end-to-end with a next-observation reconstruction loss. The size of the latent state is re-tuned in a separate hyperparameter sweep. This is the standard setup in deep RL.

\subsubsection{Model Rollouts} \label{sec:model_rollouts}

\begin{figure}[t]
  \centering
  \begin{tikzpicture}
    \node[anchor=south west,inner sep=0] (image) at (0,0) {\includegraphics[width=\linewidth]{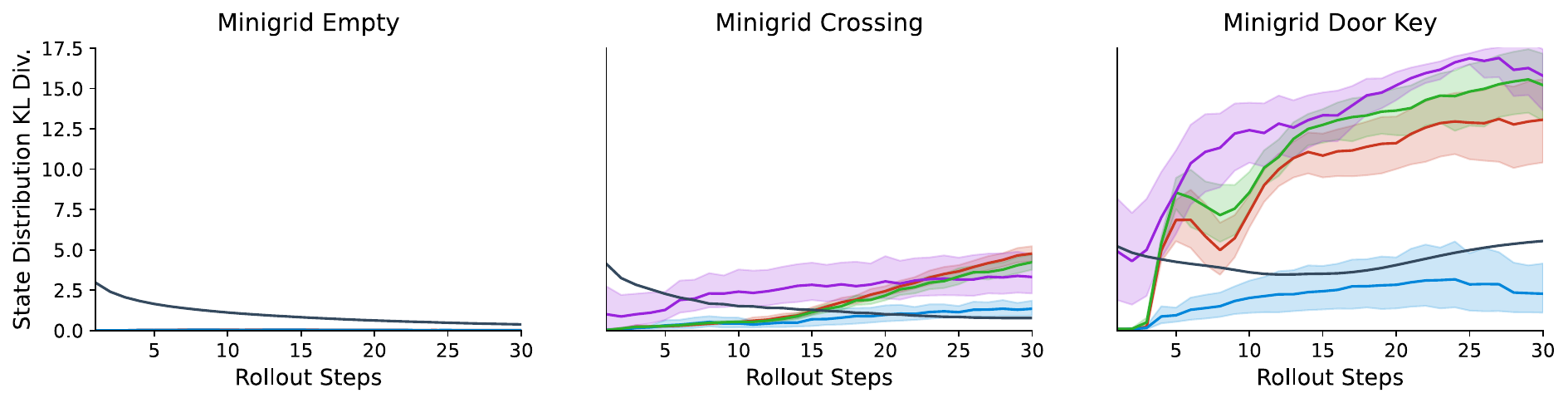}};
    \begin{scope}[x={(image.south east)},y={(image.north west)}]
      \node [anchor=east, color=plotred, font=\plotfontksmall] (cont) at (0.992, 0.54) {Vanilla AE};
      \node [anchor=east, color=plotblue, font=\plotfontksmall] (disc) at (0.62, 0.43) {VQ-VAE};
      \node [anchor=east, color=plotgreen, font=\plotfontksmall] (fta) at (0.88, 0.42) {FTA AE};
      \node [anchor=east, color=plotpurple, font=\plotfontksmall] (e2e) at (0.53, 0.38) {End-to-End};
      \node [anchor=east, color=plotnblue, font=\plotfontksmall] (uniform) at (0.34, 0.26) {Uniform};
      
      \draw [color=plotblue, line width=1pt] (disc.south) -- ++(0.06, -0.11);
      \draw [color=plotgreen, line width=1pt] (fta.west) -- ++(-0.018, 0.04);

    \node[anchor=south west, inner sep=0] (inset) at (0.085, 0.4) {%
        \adjustbox{trim={18.5} {15} {18} {15},clip}{%
            \includegraphics[width=0.25\linewidth,]{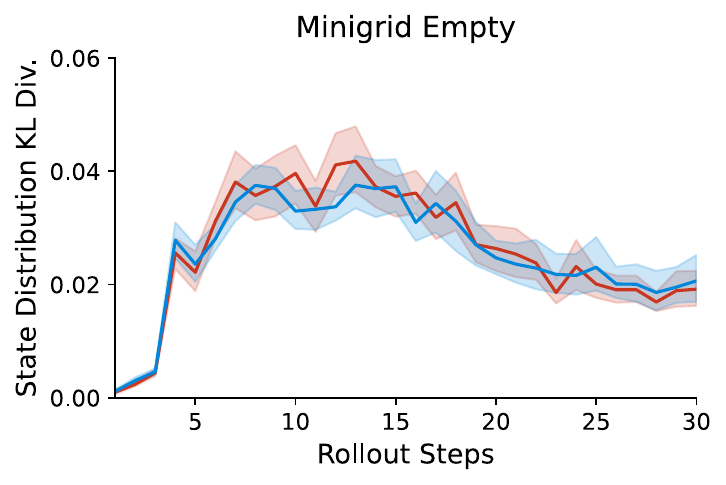}}};
    \draw (inset.south west) rectangle (inset.north east); 

      \coordinate (mainBoxSW) at (0.063, 0.183);
      \coordinate (mainBoxNE) at (0.22, 0.202);
      \draw[dashed, color=black] (mainBoxSW) rectangle (mainBoxNE);
      
      \draw[color=black, dashed] (mainBoxNE) -- (inset.south east);
      \draw[color=black, dashed] ([xshift=0cm]mainBoxNE -| mainBoxSW) -- (inset.south west);

    \end{scope}
  \end{tikzpicture}
  
  \caption{The mean KL divergence between the ground-truth and the world model induced state distributions. Lower values are better, indicating a closer imitation of the real environment dynamics. The VQ-VAE and Vanilla AE learn near-perfect models in the \textit{empty} environment, so the curves are so close to zero that they are not visible without maginification. FTA AE and End-to-End experiments were not run in the \textit{empty} environment because of the triviality. Curves depict a 95\% confidence intervals over 20 runs.}
  \label{fig:model_baseline}
\end{figure}
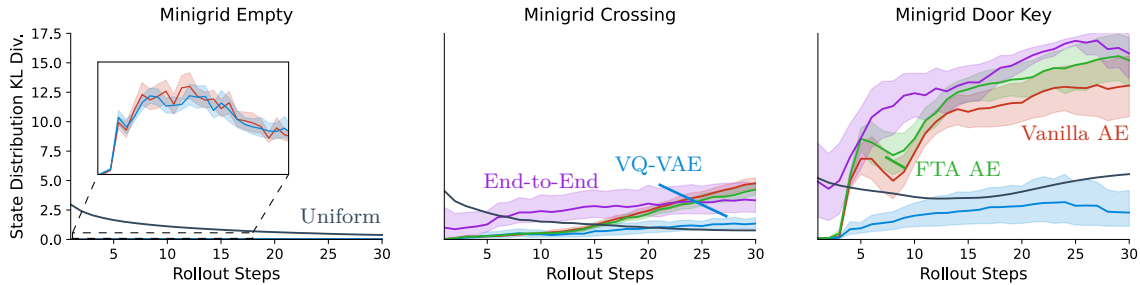

\looseness=-1
We roll out the trained world models for 30 steps and evaluate their accuracy, plotting the results in Figure \ref{fig:model_baseline}. Although all of the methods perform the same in the \textit{empty} environment, the gap in accuracy widens as the complexity progressively increases in the \textit{crossing}, and then in the \textit{door key} environment.

\looseness=-1
We examine visualizations of trajectories to better understand the patterns observed in Figure~\ref{fig:model_baseline}, showing two visualizations that most clearly represent these patterns in Figures \ref{fig:crossing_state_distrib} and \ref{fig:door_key_state_distrib} in Appendix~\ref{sec:appendix_world_model}. The trajectories predicted by the continuous models (Vanilla AE and FTA AE) in the \textit{crossing} environment rarely make it across the gap in the wall, which manifests as a steady increase in the KL divergence starting around step 14. The performance of the continuous model in the \textit{door key} environment suffers much earlier as the model struggles to predict the agent picking up the key, and again as the model struggles to predict the agent passing through the door. Notably, these two actions occur infrequently in the training data because the training data is generated with random walks, and because they can only happen once per episode even when they do occur. \textit{Stated concisely, the discrete world model more accurately predicts transitions that occur less frequently in the training data.}

\subsubsection{Scaling the World Model}

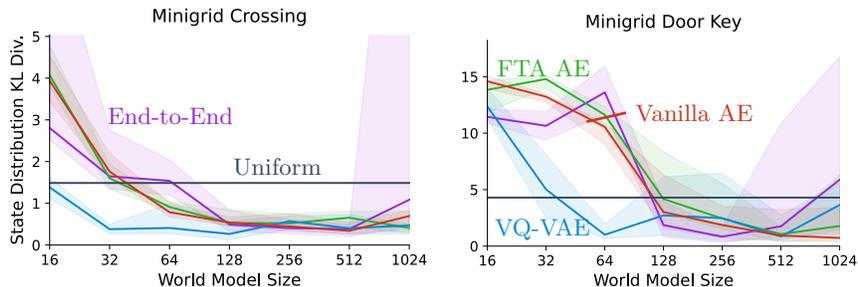
\begin{figure}[t]
  
      \centering
      \begin{tikzpicture}
        \node[anchor=south west,inner sep=0] (image) at (0,0) {
        
            \begin{subfigure}{0.38\textwidth}
            \centering
            \includegraphics[width=\linewidth]{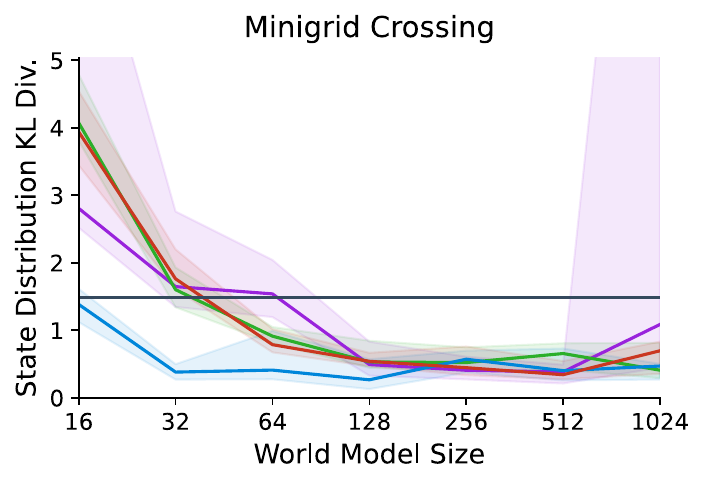}
          \end{subfigure}
          \begin{subfigure}{0.38\textwidth}
            \centering
            \adjustbox{trim={11} {0} {0} {0},clip}%
            {\includegraphics[width=\linewidth]{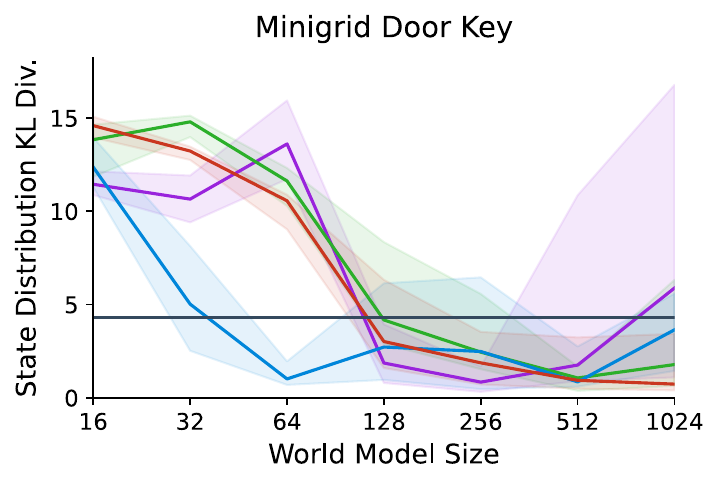}}
          \end{subfigure}
      
        };
        \begin{scope}[x={(image.south east)},y={(image.north west)}]
          \node [anchor=east, color=plotred, font=\plotfontsmall] (cont) at (0.865, 0.62) {Vanilla AE};
          \node [anchor=east, color=plotblue, font=\plotfontsmall] (disc) at (0.68, 0.23) {VQ-VAE};
          \node [anchor=east, color=plotgreen, font=\plotfontsmall] (fta) at (0.68, 0.77) {FTA AE};
          \node [anchor=east, color=plotpurple, font=\plotfontsmall] (e2e) at (0.275, 0.61) {End-to-End};
          \node [anchor=east, color=plotnblue, font=\plotfontsmall] (uniform) at (0.375, 0.44) {Uniform};
          
          \draw [color=plotred, line width=1pt] (cont.west) -- ++(-0.045, -0.03);
        \end{scope}
      \end{tikzpicture}

      \caption{\looseness=-1
      The median KL divergence between the ground-truth and the world model induced state distributions, averaged over 30 steps. Lower is better, indicating a closer imitation of the real environment dynamics. The x-axis gives the number of hidden units per layer for all three layers of the world model. The shaded region depicts a 95\% confidence interval over 20 runs. Error bars are wide for the end-to-end method due to a few divergent runs. Training the end-to-end model is harder because gradients for multiple objectives must be passed back in time through multiple steps.}
      \label{fig:model_trans_ablation}
    \end{figure}

Despite sweeping over the latent vector dimensions of the vanilla and FTA autoencoders in the hyperparameter sweep, we were unable to find an encoder architecture that enabled either of the continuous world models to adequately learn transitions underrepresented in the training data. \textbf{Either the discrete representations allow learning something that is not learnable with the continuous representations, or the fixed size of the world model is limiting the continuous model's performance.} We test the latter hypothesis by varying the size of the world model while tuning the latent dimensions of each autoencoder as described in Appendix~\ref{sec:ae_architecture}. We plot the average performance of each world model in Figure \ref{fig:model_trans_ablation}.

\looseness=-1
In the plot, an interesting pattern emerges: the performance of all methods become indistinguishable beyond a certain size of the world model. Only when the environment dynamics cannot be modeled near-perfectly, due to the limited capacity of the world model, do the discrete representations prove beneficial. As the size of the world model shrinks, the performance of the continuous models degrade more rapidly. This observation aligns with the findings in the previous section, where the performance gap between models widened with the complexity of the environment. Both results point to the same conclusion: \textit{the discrete VQ-VAE representations enable learning a more accurate world model with less modeling capacity}. This gap is notable especially when the world is much larger than what the agent has capacity to model. In this setting in our experiments, discrete representations are favorable because they allow the agent to learn more despite its limited capacity.

\subsubsection{Representation Matters} \label{sec:representation_matters}

\looseness=-1
Our goal in the previous experiments was to assess how a change in representation \textit{alone} can affect performance, but VQ-VAEs may affect more than just the representation learned. Latent spaces are defined by both the information they represent---informational content---and by the way that information is structured---representation. Because the altered bottleneck structure and objectives of a VQ-VAE may change what is learned, the previous experiments do not directly control for differences in information content. Our next experiment controls for this factor as we ask the question: \textbf{do the benefits of discrete world models stem from the representation or from the informational content of the latent states?}

\looseness=-1
To answer this question, we rerun the model learning experiment with two types of latents, both produced by the same VQ-VAE but represented in different ways. Generally, the outputs of a VQ-VAE are quantized by ``snapping'' each latent to the nearest of a finite set of embedding vectors. The resulting \textit{quantized latents} are discrete in the sense that each can take only a finite number of distinct values, but they are element-wise continuous. In our work, we alternatively represent latents as (one-hot encoded) indices of the nearest embedding vectors, which are element-wise binary. Both of these methods encode the same informational content and can produce latents of the same shape, but have different representations. If the representation of the latent space does not matter, then we would expect models learned over both representations to perform similarly.

  \begin{figure}[t]
  
      \centering
      \begin{tikzpicture}
        \node[anchor=south west,inner sep=0] (image) at (0,0) {
        
            \begin{subfigure}{0.38\textwidth}
            \centering
            \includegraphics[width=\linewidth]{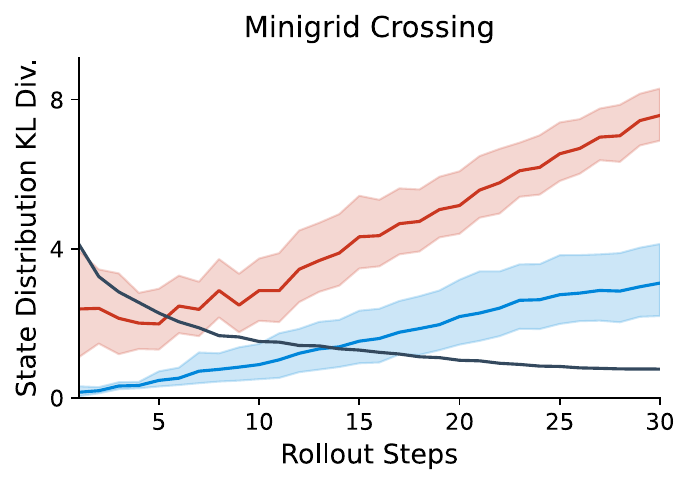}
          \end{subfigure}
          \begin{subfigure}{0.38\textwidth}
            \centering
            \adjustbox{trim={11} {0} {0} {0},clip}%
            {\includegraphics[width=\linewidth]{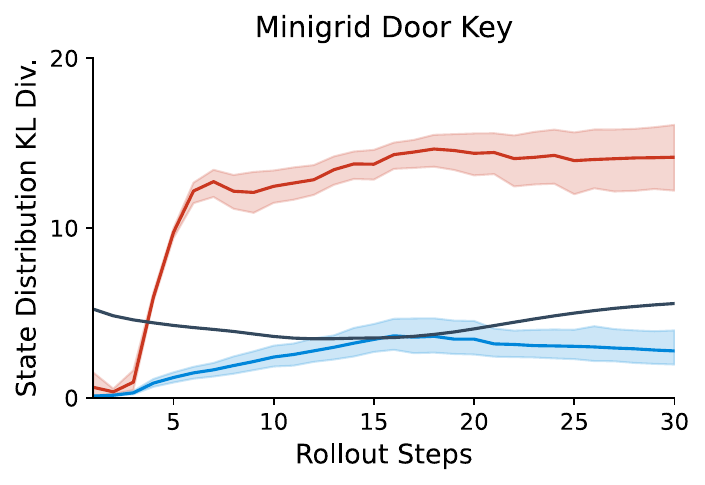}}
          \end{subfigure}
      
        };
        \begin{scope}[x={(image.south east)},y={(image.north west)}]
          \node [anchor=east, color=plotred, font=\plotfontsmall] (cont) at (0.34, 0.7) {Quantized};
          \node [anchor=east, color=plotblue, font=\plotfontsmall] (disc) at (0.47, 0.45) {Multi-One-Hot};
          \node [anchor=east, color=plotnblue, font=\plotfontsmall] (uniform) at (0.486, 0.288) {Uniform};
        \end{scope}
      \end{tikzpicture}

      \caption{
      The mean KL divergence between the ground-truth and the world model induced state distributions. Lower values are better, indicating a closer imitation of the real environment dynamics. Both methods use the same VQ-VAE architecture, but represent the information in different ways. Curves depict 95\% confidence intervals over 20 runs.}
      \label{fig:model_quantized}
    \end{figure}
    
We prepare the experiment by constructing architecturally equivalent world models with quantized and multi-one-hot representations. The number and dimensionality of the embedding vectors are set to 64 so that both representations take the same shape. The quantized model is trained with the squared error loss, but both models otherwise follow the same training procedure.

\looseness=-1
We plot the accuracy of both models in Figure \ref{fig:model_quantized}, where we see multi-one-hot representations vastly outperform quantized representations despite both being discrete and semantically equivalent. These results support the claim that \textit{the representation, rather than the informational content, is responsible for the superior performance of the VQ-VAE latents} in our experiments. Our results also suggest that \textit{the superior performance of discrete representations is not necessarily attributable to their ``discreteness'', but rather to their sparse, binary nature}. Both quantized and multi-one-hot representations are discrete and semantically equivalent, yet yield different results. These results suggest that the implicit choice of representing discrete values as multi-one-hot vectors is essential to the success of discrete representations, yet to our knowledge, such a choice is not discussed in any prior work.

\section{Model-Free RL with Discrete Representations} \label{sec: policy_learning}

We now progress to the full RL problem. Our first experiments aim to understand the effects of using discrete representations in the standard, episodic RL setting. After identifying a clear benefit, we progress to the continual RL setting with continually changing environments \citep{loss_plasticity_crl_2023} as a proxy for environments that are too big for the agent to perfectly model.

\looseness=-1
We train all RL agents in this section with the clipping version of proximal policy optimization (PPO) \citep{ppo_2017}. Instead of observations, the policy and value functions intake learned representations. Separate networks are used for the policy and value functions, but both share the same architecture: an MLP with two hidden layers of 256 units and ReLU activations. We sweep over select hyperparameters for PPO and over autoencoder hyperparameters as described in Section \ref{table:ppo_hyperparams}.

\looseness=-1
The training loop alternates between collecting data, training the actor-critic model, and training the autoencoder, as detailed in Algorithm \ref{alg:rl_training} in Appendix \ref{sec:supplemental_rl}. This setup differs from previous experiments in that environment interaction and the training of each component happen in tandem instead of in separate phases. The objectives, however, remain separate; PPO gradients only affect the policy and value function weights, and autoencoder gradients only affect the encoder. Only the end-to-end baseline is an exception, in which the entire model is trained with PPO, as is often standard in deep RL. Agents are trained in the \textit{crossing} and \textit{door key} environments shown in Figure \ref{fig:envs}. The maximum episode length is set to 400 in the \textit{crossing} environment and 1,000 in the \textit{door key} environment.

\subsection{Episodic RL}

\captionsetup[subfigure]{skip=1pt}
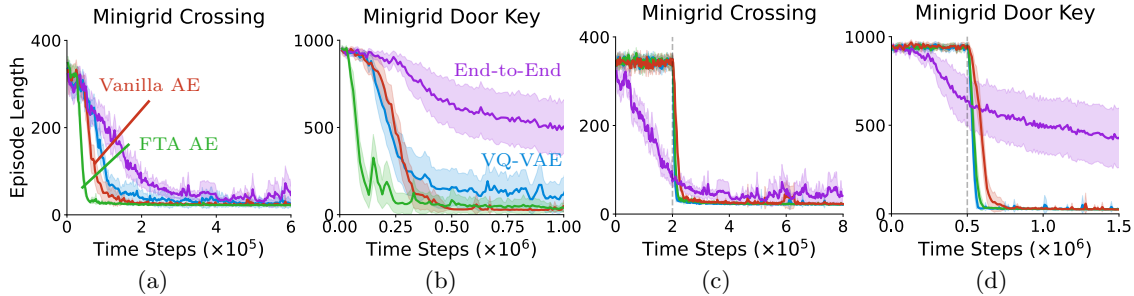
\begin{figure}[t]
  \centering
  \begin{tikzpicture}
        \node[anchor=south west,inner sep=0] (image1) at (0,0) {
          \begin{subfigure}{0.26\textwidth}
            \centering
            \includegraphics[width=\linewidth]{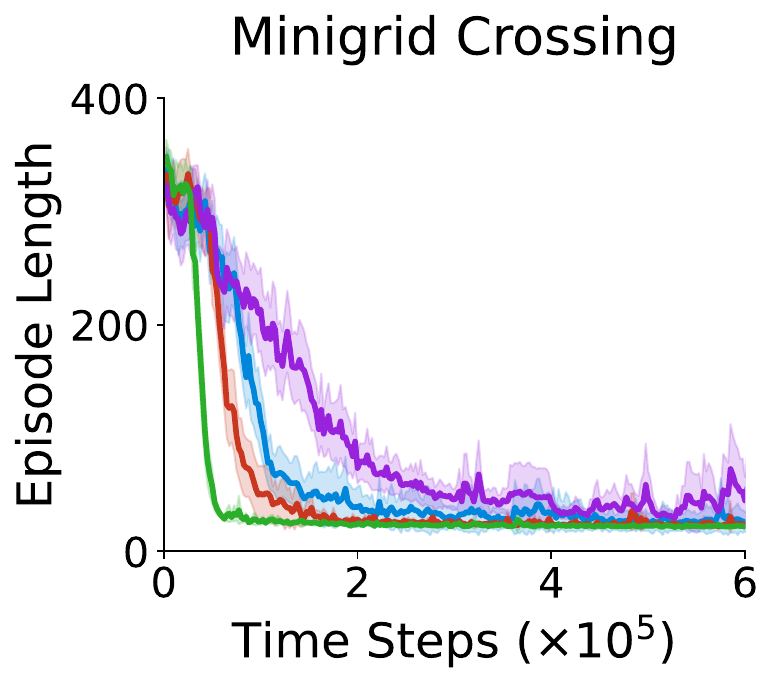}
            \caption{}
            \label{fig:vanilla_rl_perf_a}
          \end{subfigure}
        };
        \begin{scope}[x={(image1.south east)},y={(image1.north west)}]
          \node [anchor=east, color=plotred, font=\plotfontvsmall] (cont) at (0.69, 0.72) {Vanilla AE};
          \node [anchor=east, color=plotgreen, font=\plotfontvsmall] (fta) at (0.75, 0.52) {FTA AE};
          
          \draw [color=plotred, line width=1pt] (cont.south) -- ++(-0.185, -0.215); 
          \draw [color=plotgreen, line width=1pt] (fta.west) -- ++(-0.16, -0.15); 
        \end{scope}
        
        \hspace{-3mm}  
        \node[anchor=south west,inner sep=0] (image2) at (image1.south east) {
          \begin{subfigure}{0.28\textwidth}
            \centering
            \adjustbox{trim={10.5} {0} {0} {0},clip}%
            {\includegraphics[width=\linewidth]{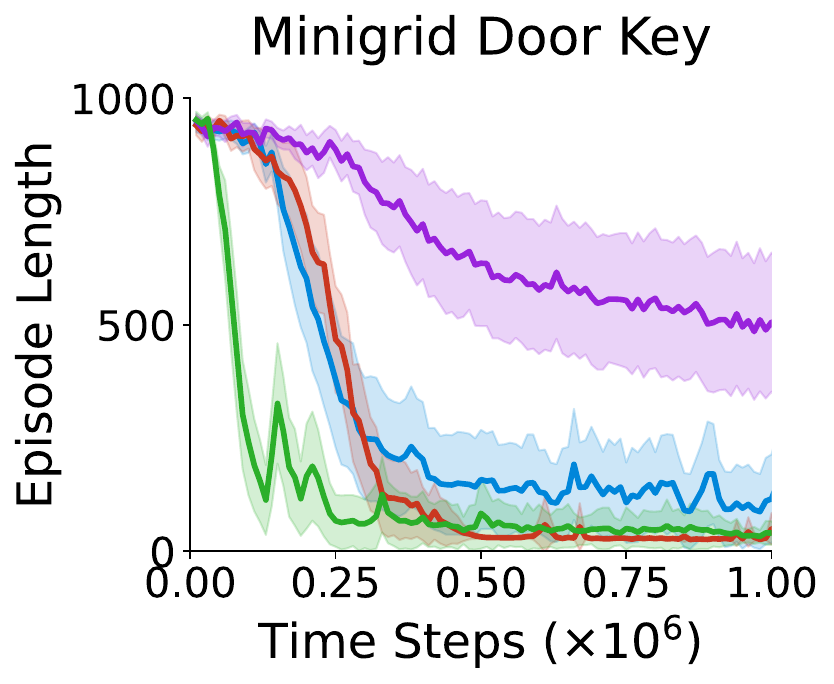}}
            \caption{}
            \label{fig:vanilla_rl_perf_b}
          \end{subfigure}
        };
        
        \begin{scope}[x={(image2.south east)},y={(image2.north west)}]
          \node [anchor=east, color=plotblue, font=\plotfontvsmall] (disc) at (0.73, 0.46) {VQ-VAE};
          \node [anchor=east, color=plotpurple, font=\plotfontvsmall] (rl) at (0.58, 0.77) {End-to-End};
          
        \end{scope}
        
        \hspace{-5mm}  
        \node[anchor=south west,inner sep=0] (image3) at (image2.south east) {
          \begin{subfigure}{0.264\textwidth}
            \centering
            \adjustbox{trim={10} {0} {0} {0},clip}%
            {\includegraphics[width=\linewidth]{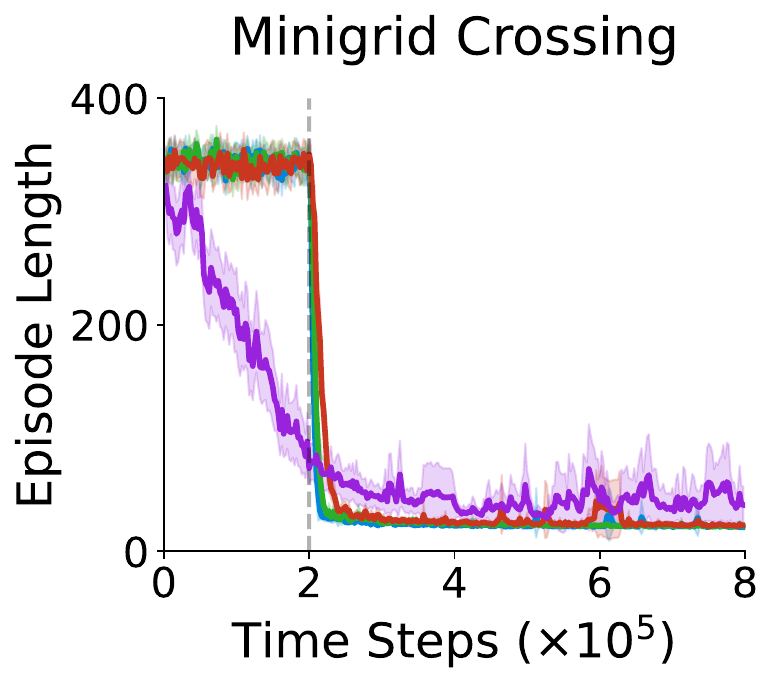}}
            \caption{}
            \label{fig:vanilla_rl_perf_c}
          \end{subfigure}
        };
        
        \hspace{-5mm}  
        \node[anchor=south west,inner sep=0] (image4) at (image3.south east) {
          \begin{subfigure}{0.28\textwidth}
            \centering
            \adjustbox{trim={10.5} {0} {0} {0},clip}%
            {\includegraphics[width=\linewidth]{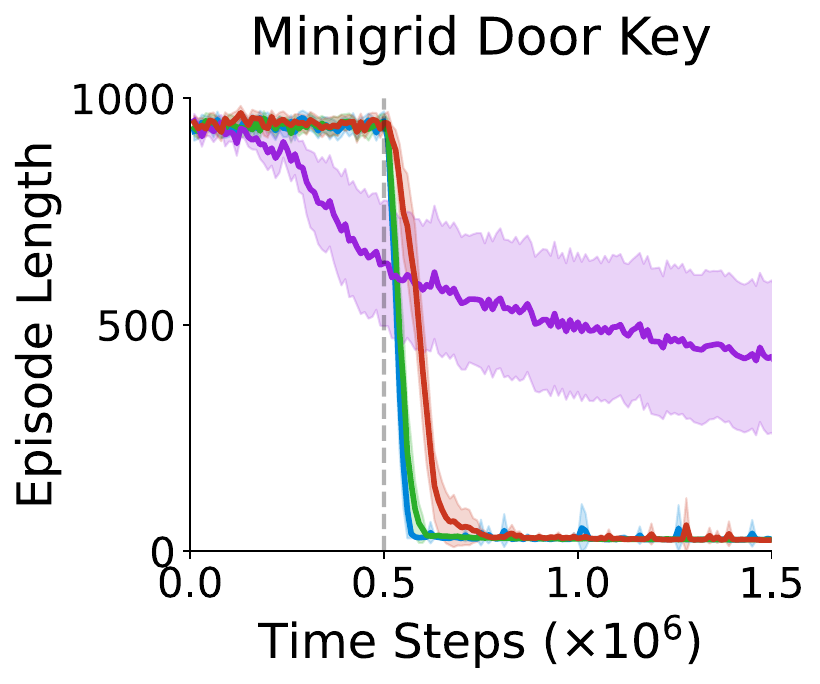}}
            \caption{}
            \label{fig:vanilla_rl_perf_d}
          \end{subfigure}
        };
  \end{tikzpicture}

  \caption{Performance of RL agents as measured by episode length with a 95\% confidence interval over 30 runs. Lower is better. (a-b) Agents are trained with PPO and autoencoder objectives from the beginning. (c-d) The PPO objective is introduced only after the dotted line (with the exception of the end-to-end method).}
  \label{fig:vanilla_rl_perf}
\end{figure}

We train RL agents with each type of representation in the \textit{crossing} and \textit{door key} environments, plotting the results in Figures \ref{fig:vanilla_rl_perf_a} and \ref{fig:vanilla_rl_perf_b}. All of the methods with an explicit representation learning objective perform better than end-to-end RL. In a reverse from the previous model learning results, the VQ-VAE now performs the worst of all the representation learning methods. Inspecting the autoencoder learning curves in Figure \ref{fig:vanilla_rl_recon} in Appendix~\ref{sec:supplemental_rl}, however, reveals an important detail: all of the autoencoders learn at different speeds. If the speed of the RL learning updates is our primary concern (whether it actually is will be discussed later), then the learning speed of the autoencoder is a confounding factor in our analysis. We address this by delaying PPO updates until all autoencoders are trained to around the same loss and plot the results in Figures \ref{fig:vanilla_rl_perf_c} and \ref{fig:vanilla_rl_perf_d}. Though the gap in performance in the new results looks small, \textit{the VQ-VAE and FTA autoencoder methods converge with around two to three times less PPO updates than the vanilla autoencoder}.

\subsection{Continual RL} \label{sec:continual_rl}

While static Minigrid environments can test these representation learning methods to an extent, they do not reflect the vastness of the real world. When the size of the world and the complexity of its problems dwarf that of the agent, the agent will lose its ability to perfectly model the world and learn perfect solutions \citep{alberta_plan_2022}. The agent must instead continually adapt in response to its limited capacity if it is to best achieve its goal(s) in this continual RL setting \citep{cl_as_constrained_rl_2023}. Given the ability of these representation learning methods to expedite policy learning, they may be well suited for the continual RL setting, where fast adaptation is key.

\begin{figure}[t]
  \centering
  \begin{tikzpicture}
    \node[anchor=south west,inner sep=0] (image1) at (0,0) {
      \begin{subfigure}{0.265\textwidth}
        \centering
        \includegraphics[width=\linewidth]{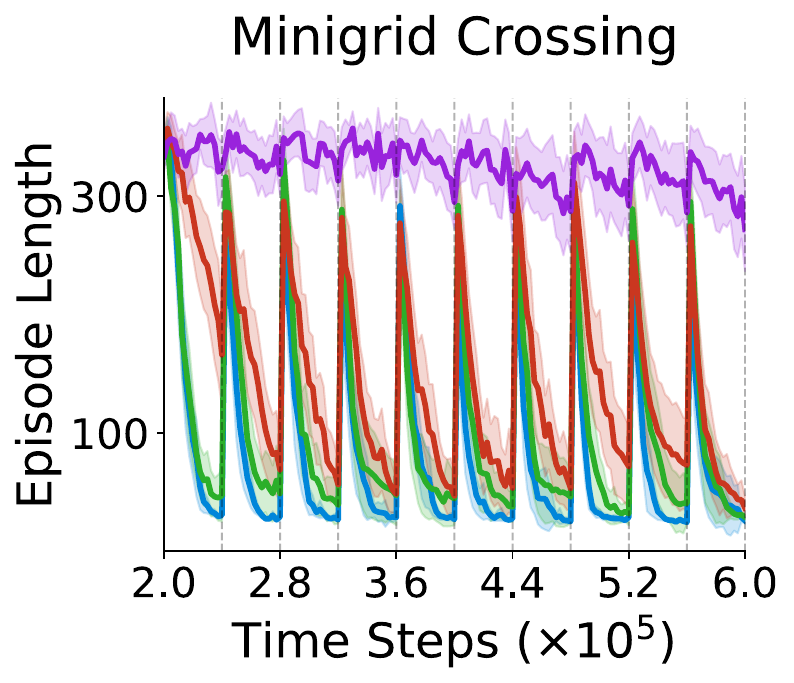}
        \caption{}
        \label{fig:changing_env_results_a}
      \end{subfigure}
    };
    
    \begin{scope}[x={(image1.south east)},y={(image1.north west)}]
      \node [anchor=east, color=plotred, font=\plotfontvsmall] (cont) at (0.56, 0.865) {Vanilla AE};
      \node [anchor=east, color=plotpurple, font=\plotfontvsmall] (rl) at (0.97, 0.865) {End-to-End};
      
      \draw [color=plotred, line width=1pt] (cont.south) -- ++(-0.12, -0.1); 
    \end{scope}
    
    \hspace{-2.8mm}  
    \node[anchor=south west,inner sep=0] (image2) at (image1.south east) {
      \begin{subfigure}{0.275\textwidth}
        \centering
        \adjustbox{trim={10.5} {0} {0} {0},clip}%
        {\includegraphics[width=\linewidth]{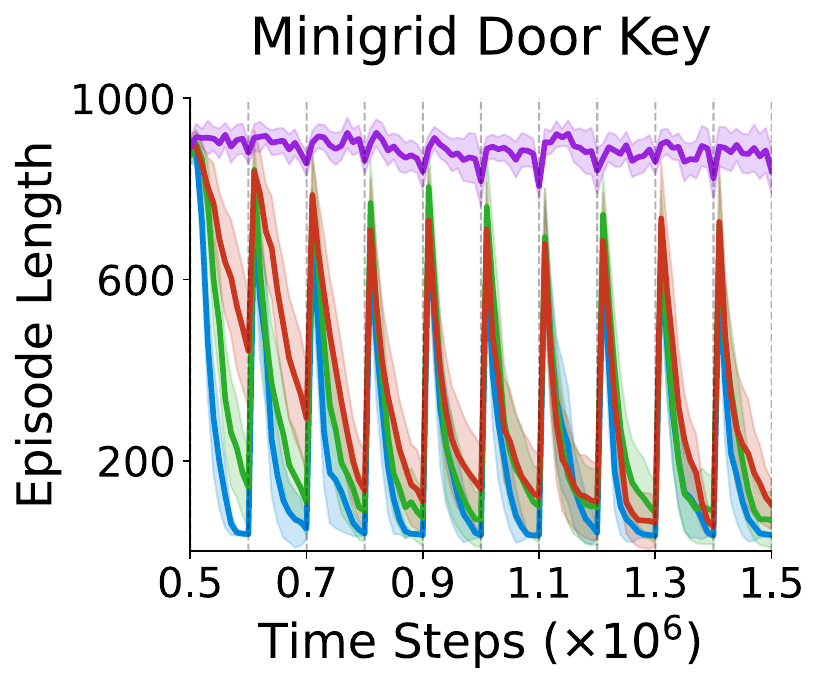}}
        \caption{}
        \label{fig:changing_env_results_b}
      \end{subfigure}
    };
    
    \hspace{-2mm}  
    \node[anchor=south west,inner sep=0] (image3) at (image2.south east) {
      \begin{subfigure}{0.259\textwidth}
        \centering
        \includegraphics[width=\linewidth]{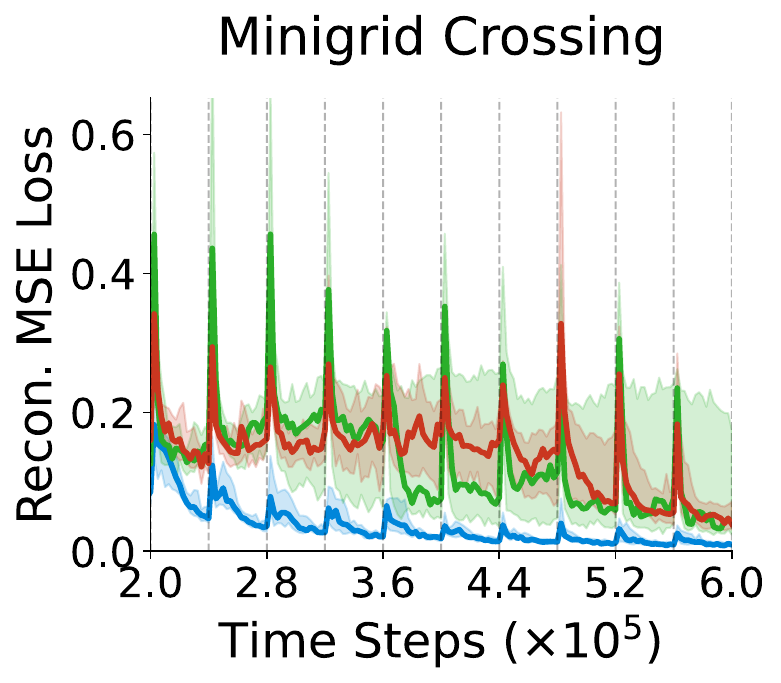}
        \caption{}
        \label{fig:changing_env_results_c}
      \end{subfigure}
    };
    
    \begin{scope}[x={(image3.south east)},y={(image3.north west)}]
      \node [anchor=east, color=plotblue, font=\plotfontvsmall] (disc) at (0.45, 0.67) {VQ-VAE};
      \draw [color=plotblue, line width=1pt] (disc.south) -- (0.56, 0.35); 
    \end{scope}
    
    \hspace{-2mm}  
    \node[anchor=south west,inner sep=0] (image4) at (image3.south east) {
      \begin{subfigure}{0.26\textwidth}
        \centering
        \adjustbox{trim={9.5} {0} {0} {0},clip}%
        {\includegraphics[width=\linewidth]{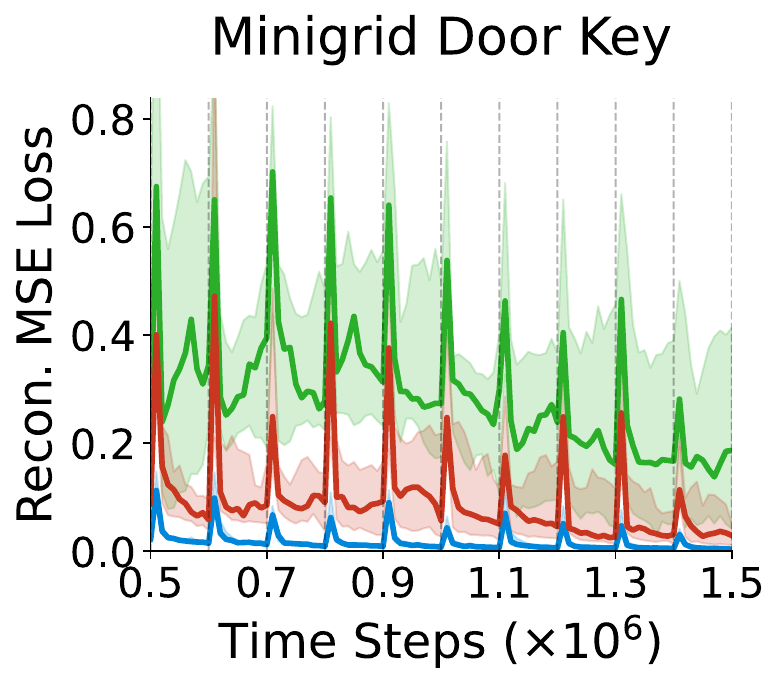}}
        \caption{}
        \label{fig:changing_env_results_d}
      \end{subfigure}
    };
    
    \begin{scope}[x={(image1.south east)},y={(image4.north west)}]
      \node [anchor=east, color=plotgreen, font=\plotfontvsmall] (fta) at (1.65, 0.7) {FTA AE};
    \end{scope}
    
  \end{tikzpicture}

  \caption{(a-b) Mean agent performance as the environments change at intervals indicated by the dotted, gray lines. Lower is better. (c-d) Median encoder reconstruction loss. Lower peaks mean the representation generalizes better, and a quicker decrease means the autoencoder is learning faster. Overall, a lower reconstruction loss is better. (a-d) Curves depict 95\% confidence intervals over 30 runs. Performance is plotted after an initial delay to learn representations, after which all methods are trained with PPO. Refer to Figure \ref{fig:full_changing_results} in Appendix \ref{sec:supplemental_rl} for the full figure.}
  \label{fig:changing_env_results}
\end{figure}
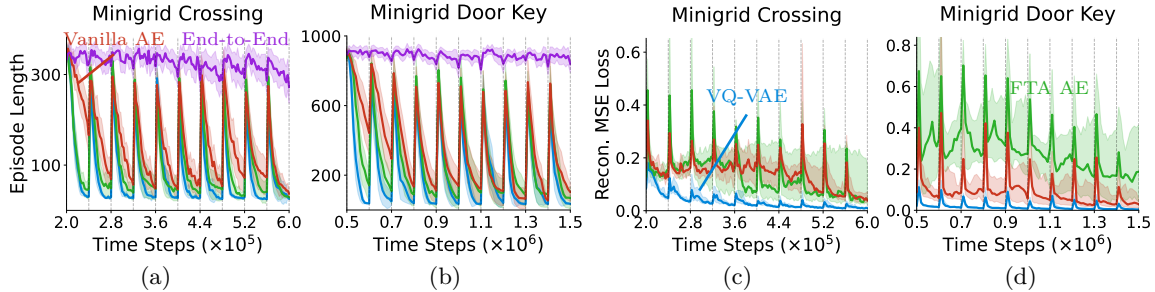

To test this hypothesis, we modify the previous experimental RL setup by randomizing the layout of the \textit{crossing} environment every 40,000 steps, and the layout of the \textit{door key} environment every 100,000 steps, as is similarly done in related work \citep{transfer_rl_survey_2009, cl_survey_2022, loss_plasticity_crl_2023}. All of the same items and walls remain, but their positions are randomized, only the positions of the goal and outer walls remaining constant. Example layouts are shown in Figure \ref{fig:changing_envs} in Appendix~\ref{sec:supplemental_rl}. By only changing the environment after a long delay, we create specific points in the learning process where we can observe the difference between how the different types of representation methods adapt to change. The RL training process otherwise stays the same, and is specified in Algorithm \ref{alg:rl_training} in Appendix~\ref{sec:supplemental_rl}. With only this modification to the environments, we rerun the previous RL experiment with a delayed PPO start, and plot the results in Figures~\ref{fig:changing_env_results_a}~and~\ref{fig:changing_env_results_b}.

\captionsetup[subfigure]{skip=1pt}

\begin{table}[H]
  \centering
  \begin{tabular}{l|c|c}
    \textbf{Latent Type} & \textbf{\textit{Crossing} Reward} & \textbf{\textit{Door Key} Reward} \\
    \hline
    End-to-End & $28 \pm 5$ & $14 \pm 2$ \\
    \hline
    Vanilla AE & $382 \pm 33$ & $866 \pm 94$ \\
    \hline
    FTA AE & $574 \pm 57$ & $1033 \pm 130$ \\
    \hline
    VQ-VAE & \boldmath{$674 \pm 21$} & \boldmath{$1324 \pm 64$} \\
  \end{tabular}
  \caption{RL performance per environment layout (95\% CI)}
  \label{table:changing_rl_reward}
\end{table}

We observe a spike in the episode length each time the environment changes, indicating that the agents' previous policies are no longer sufficient to solve the new environments. While the representation learning methods clearly outperform end-to-end training, the confidence intervals overlap at many time steps. If we instead, however, consider the average reward accumulated by each method per layout as displayed in Table \ref{table:changing_rl_reward}, a clear ranking emerges. In the \textit{crossing} environment we see $\text{VQ-VAE} > \text{FTA AE} > \text{Vanilla AE}$, and in the \textit{door key} environment we see $\text{VQ-VAE} > \text{FTA AE} \approx \text{Vanilla AE}$. 

While the slower initial learning speed of the VQ-VAE hinders its ability to maximize reward at the beginning of the training process (when PPO updates are not delayed), it does not seem to hinder its ability to adapt after an initial representation has already been learned. Inspecting the reconstruction loss of both autoencoders, plotted in Figures \ref{fig:changing_env_results_c} and \ref{fig:changing_env_results_d}, shows that the VQ-VAE's reconstruction loss increases much less when the environment changes. The shorter spikes suggest that the VQ-VAE representations generalize better, allowing them to adapt faster when the environment changes.

With these results, we return to the prior question: can multi-one-hot representations be beneficial in RL even if the initial representation is learned slower? We argue in the affirmative. If we consider continually learning RL agents in the big world setting, where the goal of the agent is to maximize reward over its lifetime by quickly adapting to unpredictable scenarios, then the cost of learning an initial representation can be amortized by a lifetime of faster adaptation.

\section{Conclusion \& Future Work}  \label{sec: conclusion}

\looseness=-1
In this work, we explored the effects of learning from discrete and sparse representations in two modules that comprise many model-based RL algorithms: model learning and model-free policy learning. When learning a world model, discrete, multi-one-hot representations enabled accurately modeling more of the world with fewer resources. When in the model-free RL setting (policy learning), agents with multi-one-hot or sparse representations learned to navigate to the goal and adapt to changes in the environment faster.

\looseness=-1
Our study underscores the advantages of multi-one-hot representations in RL but leaves several questions of deeper understanding and extrapolation to future work. We show that one-hot encoding is crucial to the success of discrete representations, but do not disentangle multi-one-hot representations from purely binary or sparse representations in our experiments. 
Prior work by \citet{fta_repr_2023} on feature generalization aligns with our results in continual RL (Section \ref{sec:continual_rl} and Appendix \ref{sec:sparsity_benefits}), and suggests that sparsity and orthogonality play a role in the success of multi-one-hot representations. Prior work on DreamerV3 \citep{dreamerv3_2023} and the success of VQ-VAEs in the domain of computer vision \citep{vqvae_2018, gen_imgs_sparse_reprs_2021, vqgan_2021, few_shot_img_gen_discrete_repr_2022} already imply that this method can extrapolate and scale to larger environments, but future work could apply these works to a wider variety of environments, beyond the inherently discrete domain of Minigrid.

Regardless of these open questions, our results implicate multi-one-hot representations learned by VQ-VAEs as a promising candidate for the representation of observations in continual RL agents. If we care about agents working in worlds much larger than themselves, we must accept that they will be incapable of perfectly representing the world. The agent will see the world as forever changing due to its limited capacity, which is the case in complex environments like the real world \citep{alberta_plan_2022, cl_as_constrained_rl_2023}. If we wish to address this issue in the representation learning space, agents must learn representations that enable quick adaptation, and are themselves quick to adapt \citep{tracking_vs_converging_2007}. The multi-one-hot representations learned in our experiments exhibit these features, and provide a potential path to build ever more efficient, continually learning RL agents.

\subsubsection*{Acknowledgments}
\label{sec:ack}
The research is supported in part by the Natural Sciences and Engineering Research Council of Canada (NSERC), the Canada CIFAR AI Chair Program, and the Digital Research Alliance of Canada. We are also grateful to Levi Lelis for his valuable feedback on an earlier version of this work. His insights led to several improvements in the paper. 

\bibliography{main}
\bibliographystyle{rlc}

\appendix

\section{Autoencoders Explained} \label{sec:ae_preliminaries}

\begin{figure}[!h]
  \centering
  \includegraphics[width=\linewidth]{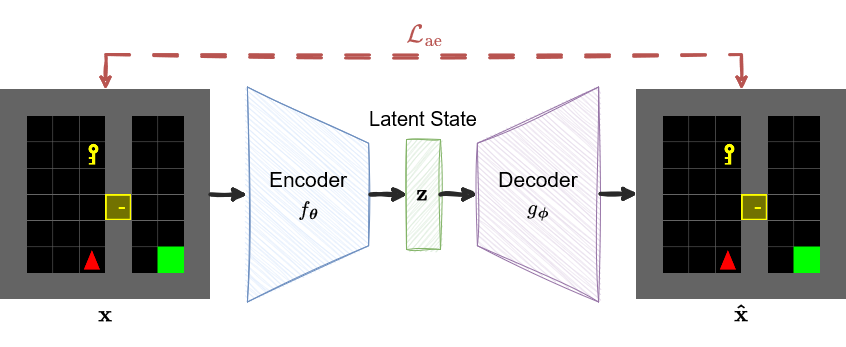}
  \caption{Depiction of a vanilla autoencoder with a continuous latent space. The input $\mathbf{x}$ is encoded with $f_{\btheta}$ to produce a latent state $\mathbf{z}$, which is decoded by $g_{\bphi}$ to produce the reconstruction $\mathbf{\hat{x}}$. The model is trained to minimize the distance between the input and reconstruction with the reconstruction loss $\mathcal{L}_{\text{ae}}$.}
  \label{fig:ae_viz}
\end{figure}

In this work, we opt to learn representations with autoencoders, neural networks with the objective of reconstructing their own inputs. Autoencoders can be decomposed into an encoder, $f_{\mathbf{\theta}}$, that projects the input into a latent space, and a decoder, $g_{\bphi}$, that attempts to reverse the transformation. Where $\mathbf{x} \in \mathbb{R}^{n}$ is an observation input to the encoder, the corresponding latent state is given by $\mathbf{z} = f_{\btheta}(\mathbf{x}) \in \mathbb{R}^{k}$, and the goal is to learn parameters $\btheta$ and $\bphi$ such that $g_{\bphi}(f_{\btheta}(\mathbf{x})) = \mathbf{x}$. We achieve this by minimizing the squared error between the input and the reconstruction over observations sampled from some dataset, $\mathcal{D}$:
\begin{equation}
\mathcal{L}_{\text{ae}} = \mathbb{E}_{\mathbf{x} \sim \mathcal{D}} \Bigl[ ||\mathbf{x} - g_{\bphi}(f_{\btheta}(\mathbf{x}))||_2^2 \Bigr] \text{.}
\label{eqn:ae_objective}
\end{equation}
Because the latent space of an autoencoder is constrained (generally by size, and sometimes by regularization), the model is encouraged to learn properties of the input distribution that are the most useful for reconstruction. We refer to this type of autoencoder, where the latent states are represented by vectors of real-valued numbers, as a vanilla autoencoder. An overview of the model is depicted in Figure \ref{fig:ae_viz}.

To learn discrete representations, we use an autoencoder variant called a vector quantized variational autoencoder (VQ-VAE) \cite{vqvae_2018}. VQ-VAEs also use an encoder, a decoder, and have the same objective of reconstructing the input, but include an additional \textit{quantization} step that is applied to the latent state between the encoder and decoder layers. After passing the input through the encoder, the resultant latent state $\mathbf{z}$ is split into $k$ latent vectors of dimension $d$: $\{\mathbf{z}_1, \mathbf{z}_2, \ldots, \mathbf{z}_k\} \in \mathbb{R}^{d}$. Each latent vector is quantized, or ``snapped'', to one of $l$ possible values specified by a set of embedding vectors. The quantization function uses $l$ embedding vectors of dimension $d$, $\{\mathbf{e}_1, \mathbf{e}_2, \ldots, \mathbf{e}_l\} \in \mathbb{R}^{d}$, which are learned parameters of the VQ-VAE.

\begin{figure}[!h]
  \centering
  \includegraphics[width=\linewidth]{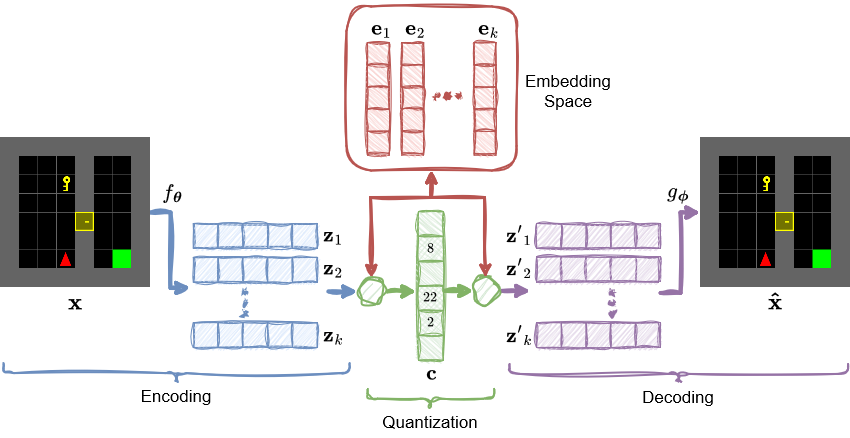}
  \caption{Depiction of the VQ-VAE architecture. The input $\mathbf{x}$ is encoded with encoder $f_{\btheta}$ to produce latent vectors $\{\mathbf{z}_1, \mathbf{z}_2, \ldots, \mathbf{z}_k\} \in \mathbb{R}^{d}$. In the first green circle, each latent vector is compared to every embedding vector to produce codebook $\mathbf{c}$, a vector of indices indicating the most similar embedding vectors (example values are depicted). In the second green circle, the indices are transformed into their corresponding embedding vectors to produce quantized vectors $\{\mathbf{z'}_1, \mathbf{z'}_2, \ldots, \mathbf{z'}_k\} \in \mathbb{R}^{d}$. The quantized vectors are then decoded by $g_{\bphi}$ to produce the reconstruction $\mathbf{\hat{x}}$. Our work uses one-hot encodings of the codebook $\mathbf{c}$ as discrete representations.}
  \label{fig:vqvae_viz}
\end{figure}

The quantization happens in two phases. First, each latent vector is compared to every embedding vector using the L2 norm, and indices of the most similar embedding vectors are returned:
\begin{equation}
c_i = \argmin_{j} \| \mathbf{z}_i - \mathbf{e}_j \|_2, \text{ for all } i = 1, 2, ...,k \text{.}
\end{equation}
The resultant vector of integers $\mathbf{c}$ is called the \textit{codebook}, and indicates which embedding vectors are the most similar to each latent vector. In the second phase, the indices in the codebook are used to retrieve their corresponding embeddings, producing the quantized latent vectors:
\begin{equation}
\mathbf{z'}_i = \mathbf{e}_{c_i}, \text{ for all } i = 1, 2, ...,k \text{.}
\end{equation}
The quantized vectors $\{\mathbf{z'}_1, \mathbf{z'}_2, \ldots, \mathbf{z'}_k\} \in \mathbb{R}^{d}$ are the final output of the quantization function, and are concatenated before being passed to the decoder. The full architecture is depicted in Figure \ref{fig:vqvae_viz}.

Because the quantization process is not differentiable, a \textit{commitment loss} is added to pulls pairs of latent states and their matching embeddings towards each other. If latent vectors are always near an existing embedding, then there will be minimal difference between all $\mathbf{z}_i$ and $\mathbf{z'}_i$, and we can use the straight-through gradients trick \cite{straight_through_2013} to pass gradients directly back from $\mathbf{z'}$ to $\mathbf{z}$ with no changes. Combining the reconstruction and commitment losses, the full objective is given by the minimization of
\begin{equation}
\mathcal{L}_{\text{vqvae}} = \mathbb{E}_{\mathbf{x} \sim \mathcal{D}} \left[ ||\mathbf{x} - g_{\bphi}(q_{\mathbf{e}}(f_{\btheta}(\mathbf{x})))||_2^2 + \beta \sum_{i=1}^{k} \| \mathbf{z}_i - \mathbf{e}_{\mathbf{z}_i} \|_2^2 \right] \text{,}
\label{eqn:vqvae_objective}
\end{equation}
where $q_{\mathbf{e}}$ is the quantization function, $\beta$ is a hyperparameter that weights the commitment loss, and $\mathbf{e}_{\mathbf{z}_i}$ is the closest embedding vector to $\mathbf{z}_i$. In practice, the speed at which the encoder weights and embedding vectors change are modified separately by weighting the gradients of both modules individually. We use a value of $\beta = 1$ in our work, and scale the embedding updates with a weight of 0.25.

The discrete representations we use for downstream tasks RL tasks are different from the quantized vectors that are passed to the decoder. We instead use one-hot encodings of the values in the codebook:
\begin{equation}
o_{ij} = 
\begin{cases} 
1 & \text{if } j = c_i, \\
0 & \text{otherwise}
\end{cases}
\quad \text{for } j = 1, 2, \dots, l.
\end{equation}
The result is a series of one-hot vectors $\{\mathbf{o}_1, \mathbf{o}_2, \ldots, \mathbf{o}_k\} \in \mathbb{R}^{l}$ that represent a single state, which we refer to as a multi-one-hot encoding or discrete representation.

\section{Stochastic World Models} \label{sec:stoch_world_models}

We use a variant of the method proposed by \citet{stochastic_muzero_2022} to learn sample models for stochastic environments. The method works similarly to a distribution model, first learning a distribution over possible outcomes during training, and then sampling from that distribution during evaluation. The problem faced by most distribution models is how to represent a distribution over a complex state space (or latent space in our case). \citeauthor{stochastic_muzero_2022} circumvent this problem by learning an encoder $e$ that discretizes each state-action pair, mapping it to a single, $k$-dimensional one-hot vector we call the outcome vector. Each of the possible $k$ values represents a different outcome of the transition.

The high-level idea is that while directly learning a distribution over full latent states is intractable, learning a categorical distribution over a limited, discrete set of outcomes (the outcome distribution) is possible. Whenever we wish to use the world model, we can sample from the outcome distribution and include the one-hot outcome vector as an additional input to the world model, indicating which of the $k$ outcomes it should produce. Table \ref{table:stoch_hyperparameters} in provides the relevant hyperparameters for this method.

\begin{table}[H]
  \centering
  \begin{tabular}{l|l}
    \textbf{Hyperparameter} & \textbf{Value} \\
    \hline
    Bin count & 32 \\
    \hline
    Discretization projection & 256, 256 \\
    \hline
    Prediction projection & 256, 256 \\
  \end{tabular}
  \caption{Stochastic sample model hyperparameters} 
  \label{table:stoch_hyperparameters}
\end{table}

\section{Autoencoder Architecture} \label{sec:ae_architecture}

The vanilla autoencoder, FTA autoencoder, and VQ-VAE use the same encoder and decoder architecture, only differing in the layer that produces the latent state. The decoder is a mirror of the encoder, reversing each of the shape transformation, so we describe only the encoder architecture. The encoder starts with three convolutional layers with square filters of sizes $\{8, 6, 4\}$, channel of sizes $\{64, 128, 64\}$, strides of $\{2, 2, 2\}$ (or $\{2, 1, 2\}$ for the crossing environment), and uniform padding of $\{1, 0, 0\}$. Each convolutional layer is followed by a ReLU activation. The downscaling convolutions are followed by an adaptive pooling layer that transforms features into a shape of $(k \times k \times 64)$, and finally a residual block \citep{resnet_2016} consisting of a convolutional layer, batch norm \citep{batch_norm_2015}, ReLU, convolutional layer, and another batch norm. These general layers are followed by layers specific to the type of autoencoder.

The vanilla autoencoder flattens the convolutional output and projects it to a latent space of size $D$ with a linear layer. We use a value of $k=8$ and sweep over values of $d=\{16, 64, 256, 1024\}$ for each environment. We use $d=64$ for the \textit{empty} environment, $d=256$ for \textit{crossing}, and $d=1024$ for \textit{door key}, though we note that we do not observe a statistically significant difference in performance for values of $d \geq 64$. The end-to-end baseline uses the same architecture and tuning procedure, but the final hyperparameter values are $d=64$ for \textit{crossing}, and $d=1024$ for \textit{door key}.

The FTA autoencoder has the same structure as the vanilla autoencoder, but with an FTA after the final bottleneck layer. The tiling bounds are fixed at $[-2, 2]$ for all cases, except for learning a world model in the \textit{door key} environment, where it is $[-4, 4]$. We sweep over values of $d=\{64, 256, 1024\}$ and the number of tiles, $k=\{8, 16, 32\}$. The sparsity parameters, $\eta$, is set to be the same as the size of the tiles, as is recommended in the original work \citep{fta_2021}. We use values of $d=64$ and $k=16$ in both environments.

The VQ-VAE directly quantizes the output of the general layers, so the only other parameters added are the embedding vectors. The number of vectors that make up a latent state is given by $k^2$, and we let $l$ be the number of embedding vectors, resulting in discrete representations of shape $(k^2, l)$. We sweep over values of $k=\{3, 6, 9\}$ and $l=\{16, 64, 256, 1024\}$ for each environment. We use $k=6$ and $l=1024$ (for a total size of 6,144) for all environments except for \textit{crossing}, which uses a value of $k=9$ (for a total size of 9,216).

When designing the experiments, we considered how to construct a fair comparison between the continuous and discrete methods despite the fact that each have different ideal sizes of the latent state, which makes one model bigger than the other. This is a particularly difficult question because it is unclear if we should focus on the size of a representation in bits, or the size of the representation in the number of values used to represent it in a deep learning system. A discrete representation is orders of magnitude smaller than a continuous representation if represented in bits ($9 \times \log_{2}{1024} = 90$ bits in the \textit{crossing} environment), but takes an order of magnitude more values to represent as one-hot vectors being passed to a neural network ($9 \times 1024 = 9216$ values in the \textit{crossing} environment). Ultimately, we found that answering this question was unnecessary,  as the performance of both methods was limited no matter how large we made the size of the representations. In the \textit{crossing} environment, for example, the performance of the continuous model would not increase even if we increased the size of the latent state from 256 to 9,216 values to match that of the discrete latent state.

\section{Reinforcement Learning Hyperparameters} \label{sec:rl_hyperparams}

Before running the model-free RL experiments, we performed a grid search over the most sensitive PPO hyperparameters for the continuous model. We swept over clipping values, $\epsilon \in \{0.1, 0.2, 0.3\}$, and the number of training epochs per batch, $n \in \{10, 20, 30, 40\}$. We use the same final PPO hyperparameters for training the RL models with FTA and VQ-VAE latents, which are provided in table \ref{table:ppo_hyperparams}.

After the sweep over PPO hyperparameters, we also repeated a sweep over the latent dimensions of all of the autoencoders (with the exception of the VQ-VAE, which we found to be robust to a large range of hyperparamers) as described in Section \ref{sec:ae_architecture}. The vanilla autoencoder and end-to-end baseline use a $d=256$ dimensional latent space. The FTA autoencoder also uses $d=256$ dimensional pre-activation latent space with $k=8$ tiles, forming a 2048-dimensional post-activation latent space. The VQ-VAE uses $k^2=36$ latent vectors and $l=256$  embedding vectors, forming a 9216-dimensional latent space.

\begin{table}[H]
  \centering\begin{tabular}{l|l}
    \textbf{Hyperparameter} & \textbf{Value} \\
    \hline
    Horizon (T) & 256 \\
    \hline
    Adam step size & 256 \\
    \hline
    (PPO) Num. epochs & 10 \\
    \hline
    (PPO) Minibatch size & 64 \\
    \hline
    Clipping value ($\epsilon$) & 0.2 \\
    \hline
    Discount ($\gamma$) & 0.99 \\
    \hline
    (Autoencoder) Num. epochs & 8
  \end{tabular}
  \caption{RL training hyperparameters} 
  \label{table:ppo_hyperparams}
\end{table}

\section{Experiment Details} \label{sec:experiment_details}

\begin{table}[H]
  \centering
  \begin{tabular}{c|c|c|c|c}
    \textbf{\begin{tabular}[c]{@{}c@{}}Environment\\Name\end{tabular}} & \textbf{\begin{tabular}[c]{@{}c@{}}Image\\Dimensions\end{tabular}} & \textbf{Actions} & \textbf{Stochastic} & \textbf{\begin{tabular}[c]{@{}c@{}}\# of\\Unique\\States\end{tabular}} \\
    \hline
    \textit{Empty} & $48 \times 48 \times 3$ & left, right, forward & no & 64\\
    \hline
    \textit{Crossing} & $54 \times 54 \times 3$ & left, right, forward & yes & 172\\
    \hline
    \textit{Door Key} & $64 \times 64 \times 3$ & \makecell{left, right, forward,\\pickup, use} & yes & 292\\
  \end{tabular}
  \caption{Minigrid environment specifications} 
  \label{table:env_stats}
\end{table}

\section{Measuring Sparsity} \label{sec:sparsity_benefits}

In Section \ref{sec:representation_matters}, our comparison between multi-one-hot and quantized VQ-VAE representations (Figure \ref{fig:model_quantized}) resulted in a decisive victory for multi-one-hot representations, which are both sparse and binary. Then in the continual RL setting in Section \ref{sec:continual_rl}, we again see the two sparse representations perform the best. These results suggest that there is an advantage to using sparse representations, but \textit{can we measure the effects of different levels of sparsity?}

In this section, we design an experiment that measures the effects of varying levels of sparsity in the continual RL setting. The most straightforward way to design such an experiment with a VQ-VAE is to change the size of the codebook, which directly controls the level of sparsity. Changing only the codebook, however, also changes the number of the parameters in the model. If we want to measure the effects of \textit{only} sparsity, then we need to control for the size of the model.

\begin{figure}[t]
  \centering
  
    \begin{subfigure}{0.49\textwidth}
    \centering
    \includegraphics[width=\linewidth]{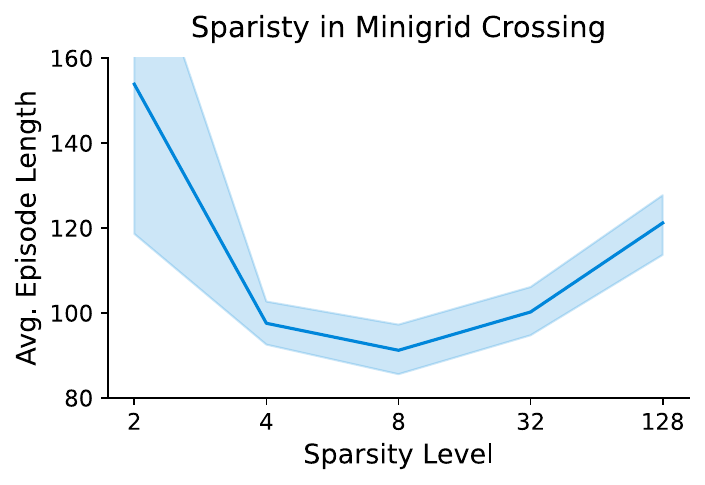}
    \end{subfigure}
    \begin{subfigure}{0.50\textwidth}
    \centering
    \adjustbox{trim={13} {0} {0} {0},clip}%
    {\includegraphics[width=\linewidth]{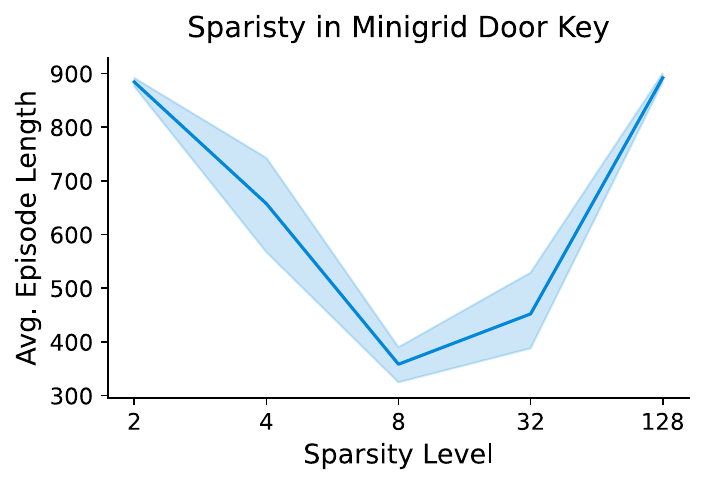}} 
    \end{subfigure}
    
    \caption{Episode length of a continual RL agent averaged over 15 runs per data point. Lower is better, indicating faster navigation to the goal. All agents use VQ-VAE representations, and the sparsity level indicates the ratio of 0s to 1s in the representation (e.g. a sparsity level of 8 indicates that there are 7 zeros for each one). The shaded region depicts a 95\% confidence interval.}
    \label{fig:vqvae_sparsity}
\end{figure}

In this experiment, we vary the dimensionality of the embeddings, the number of latents, and the size of the codebook all in tandem so that the size of the model stays constant as the level of sparsity changes. At each level of sparsity, we rerun the continual RL experiments as described in Section \ref{sec:continual_rl} and plot a summary of the results in Figure \ref{fig:vqvae_sparsity}. In the results, we see that \textbf{sparsity does help and that there is an ideal amount of sparsity}. In both the \textit{crossing} and \textit{door key} environments, a sparsity level of 8 leads to optimal performance.\footnote{Note that the optimal sparsity levels in this experiment do not align with experiments in previous sections because we use a modified architecture that allows us to change the sparsity level more freely.} These results mirror findings from the work on FTA by \citet{fta_2021}, which also show sparsity helping up to a certain threshold. 

\clearpage
\section{Supplemental World-Model Materials}
\label{sec:appendix_world_model}

\begin{figure}[H]
  \centering
  \includegraphics[width=\linewidth]{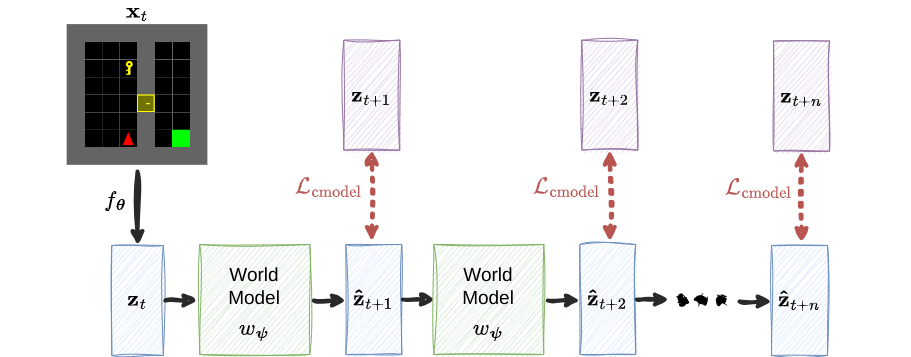}
  \caption{Depiction of a continuous world model training with $n$ steps of hallucinated replay. After encoding the initial observation, the world model rolls out a trajectory of predicted latent states, $\mathbf{\hat{z}}_{t+1}, \mathbf{\hat{z}}_{t+2}, \dotsc, \mathbf{\hat{z}}_{t+n}$. Actions from a real trajectory are used during training, but are excluded in the depiction to avoid clutter. The loss at each time step is calculated as the mean squared error between the hallucinated latent state $\mathbf{\hat{z}}_{t+i}$ and the ground-truth, $\mathbf{z}_{t+i}$. This method is called hallucinated replay because the entire trajectory after the first latent state is hallucinated by the world model.}
  \label{fig:cont_model_viz}
\end{figure}

\begin{figure}[H]
  \centering
  \includegraphics[width=\linewidth]{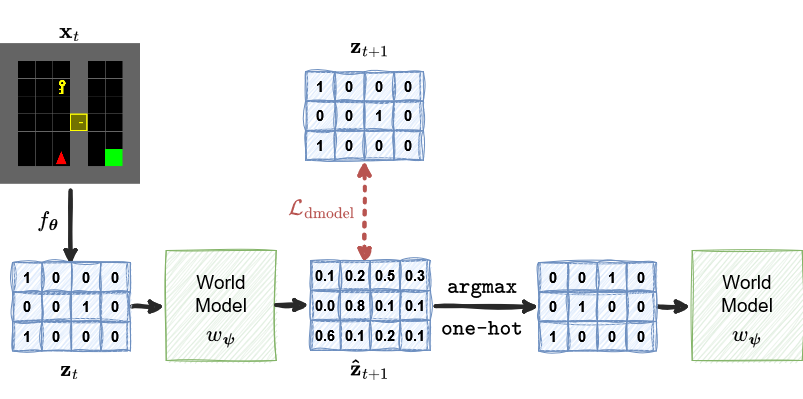}
  \caption{Depiction of a single step of discrete world model training and the subsequent discretization of the latent state. The observation $\mathbf{x}_t$ is encoded to produce latent state $\mathbf{z}_t$, which is passed to the world model to sample the logits $\mathbf{\hat{z}}_{t+1}$ for a following state. The predicted next state logits $\mathbf{\hat{z}}_{t+1}$ are compared to the ground truth state $\mathbf{z}_{t+1}$, which is constructed from the corresponding ground-truth observation: $\mathbf{z}_{t+1} = f_{\btheta}(\mathbf{x}_{t+1})$. Before the world model can be reapplied, the latent state logits must be discretized with an \texttt{argmax} operator and converted to the one-hot format.}
  \label{fig:disc_model_viz}
\end{figure}

\begin{figure}[!h]
    \centering
    \resizebox{\linewidth}{!}{
    \begin{tikzpicture}
        \def\vertspacing{0.2cm}
        
        \node[anchor=south west, inner sep=0] (img1) at (0,0) {\includegraphics{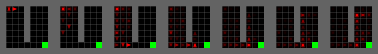}};
        \node[anchor=north west, inner sep=0] (img2) at ([yshift=-\vertspacing]img1.south west) {\includegraphics{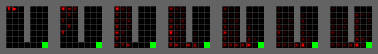}};
        \node[anchor=north west, inner sep=0] (img3) at ([yshift=-\vertspacing]img2.south west) {\includegraphics{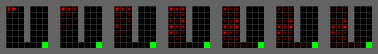}};

        \node[rotate=90, anchor=center, yshift=0.5cm] at (img1.west) {\makecell{Ground\\Truth}};
        \node[rotate=90, anchor=center, yshift=0.3cm] at (img2.west) {VQ-VAE};
        \node[rotate=90, anchor=center, yshift=0.3cm] at (img3.west) {Vanilla AE};
        
        \foreach \i/\x in {1/1, 2/5, 3/10, 4/15, 5/20, 6/25, 7/30} {
            \node[anchor=north, xshift=-0.54cm] at ($ (img3.south west) + (\i*1.905 - 0.4, 0) $) {\x};
        }

    \end{tikzpicture}
    }
    \caption{Comparison of rollouts predicted by different world models in the \textit{crossing} environment. Each row visualizes the state distributions throughout rollouts predicted by different world models, with the x-axis giving the step in the rollout. The ground-truth row depicts the state distribution over rollouts as a policy that explores the right side of the environment is enacted in the true environment. Predicted observations are averaged over 10,000 rollouts. Being closer to the ground-truth indicates a higher accuracy.}
    \label{fig:crossing_state_distrib}
\end{figure}
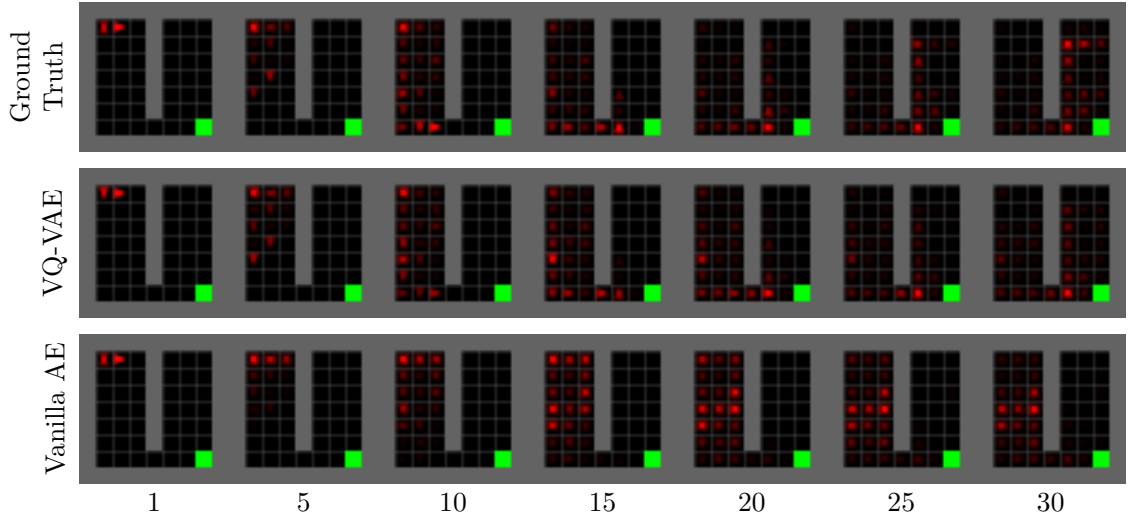

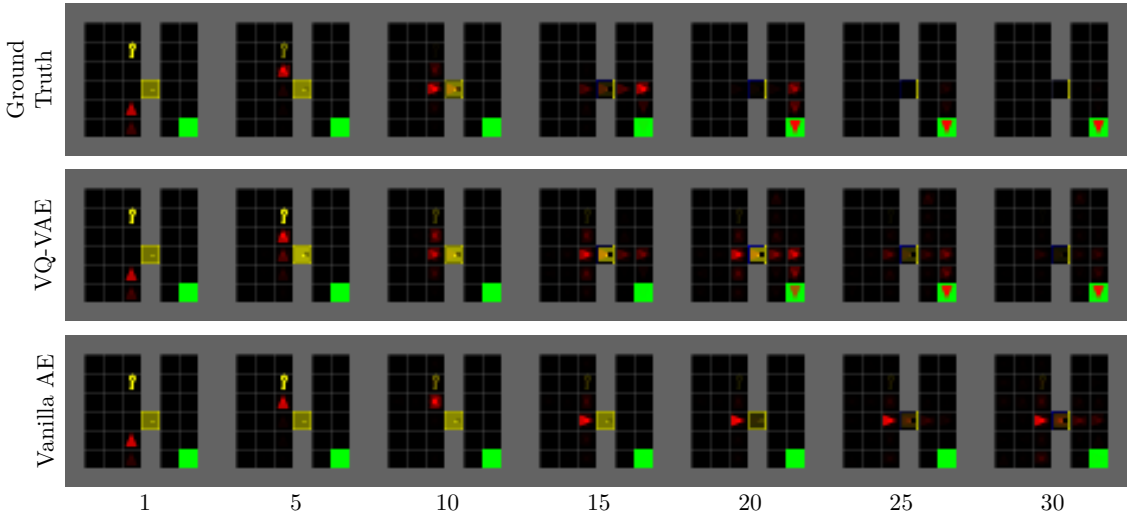
\begin{figure}[!h]
    \centering
    \resizebox{\linewidth}{!}{
    \begin{tikzpicture}
        \def\vertspacing{0.2cm}
        
        \node[anchor=south west, inner sep=0] (img1) at (0,0) {\includegraphics{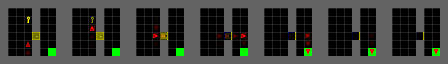}};
        \node[anchor=north west, inner sep=0] (img2) at ([yshift=-\vertspacing]img1.south west) {\includegraphics{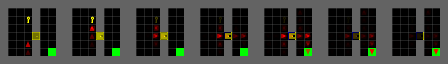}};
        \node[anchor=north west, inner sep=0] (img3) at ([yshift=-\vertspacing]img2.south west) {\includegraphics{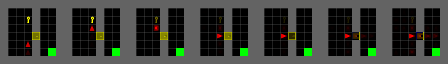}};

        \node[rotate=90, anchor=center, yshift=0.5cm] at (img1.west) {\makecell{Ground\\Truth}};
        \node[rotate=90, anchor=center, yshift=0.3cm] at (img2.west) {VQ-VAE};
        \node[rotate=90, anchor=center, yshift=0.3cm] at (img3.west) {Vanilla AE};
        
        \foreach \i/\x in {1/1, 2/5, 3/10, 4/15, 5/20, 6/25, 7/30} {
            \node[anchor=north, xshift=-0.54cm] at ($ (img3.south west) + (\i*2.255 - 0.53, 0) $) {\x};
        }

    \end{tikzpicture}
    }
    \caption{Comparison of rollouts predicted by different world models in the \textit{door key} environment. Each row visualizes the state distributions throughout rollouts predicted by different world models, with the x-axis giving the step in the rollout. The ground-truth row depicts the state distribution over rollouts as a policy that navigates to the goal state is enacted in the true environment. Predicted observations are averaged over 10,000 rollouts. Being closer to the ground-truth indicates a higher accuracy.}
    \label{fig:door_key_state_distrib}
\end{figure}

This section contains additional materials that help describe the model training process and results. Algorithm \ref{alg:world_model} provides pseudo-code for the training algorithm, Figures \ref{fig:cont_model_viz} \& \ref{fig:disc_model_viz} visualize the training process, and Figures \ref{fig:crossing_state_distrib} \& \ref{fig:door_key_state_distrib} visualize distributions of rollouts predicted by the learned world models.

\begin{algorithm}
\caption{Training Autoencoder and World Model} \label{alg:world_model}
\begin{algorithmic}

\STATE $\mathcal{D} \leftarrow \text{dataset of transition tuples } (s, a, s')$
\STATE Initialize the encoder, $f_{\btheta}$, decoder, $g_{\bphi}$, and world model, $w_{\bpsi}$
\STATE Set the number of autoencoder training steps, $N$, the number of of world model training steps, $L$, and the number of hallucinated replay steps, $K$
\\
\\
\COMMENT{Training the Autoencoder}
\FOR{$N$ steps}
    \STATE Sample transition $(s_0, a_0, s_1) \in \mathcal{D}$
    \STATE  $\mathbf{z} \leftarrow f_{\btheta}(s_0)$
    \STATE $\hat{s}_0 \leftarrow g_{\bphi}(\mathbf{z}_0)$
    \STATE loss $\leftarrow \text{MSE}(s_0, \hat{s}_0)$
    \STATE Update parameters $\btheta$ and $\bphi$ with Adam
\ENDFOR
\STATE Freeze autoencoder model weights, $\btheta$ and $\bphi$ 
\\
\\
\COMMENT{Training the World Model}
\FOR{$L$ steps}
    \STATE Sample a sequence of transitions $(s_0, a_0, s_1, a_1, ..., s_K) \in \mathcal{D}$
    \STATE $\mathbf{\hat{z}} \leftarrow f_{\btheta}(s_0)$
    \FOR{$k$ in $\{0, 1, ..., K-1\}$}
        \STATE $\mathbf{\hat{z}} \leftarrow w_{\bpsi}(\mathbf{\hat{z}}, a_k)$
        \STATE $\mathbf{z}_{k+1} \leftarrow f_{\btheta}(s_{k+1})$
        \STATE Compute loss between $\mathbf{\hat{z}}$ and $\mathbf{z}_{k+1}$ \COMMENT{cross-entropy for discrete, MSE for continuous}
        \STATE Update parameters $\bpsi$ with Adam
    \ENDFOR
\ENDFOR

\end{algorithmic}
\end{algorithm}

\clearpage

\clearpage

\clearpage
\section{Supplemental RL Materials}
\label{sec:supplemental_rl}

\begin{figure}[!hbt]
              \centering
                \begin{subfigure}{0.99\textwidth}
                \centering
                \includegraphics[width=\linewidth]{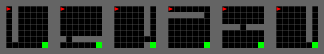}
              \end{subfigure}
              \vskip 0.2cm
              \begin{subfigure}{0.99\textwidth}
                \centering
            \includegraphics[width=\linewidth]{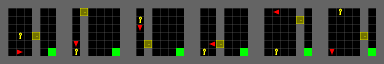}
              \end{subfigure}
              \caption{The top row depicts random initializations of the \textit{crossing} environment, and the bottom that of the \textit{door key} environment. Each time the environment changes, the positions of all internal walls and objects are randomized, with the exception of the agent position in the \textit{crossing} environment and the goal in both environments.}
              \label{fig:changing_envs}
            \end{figure}

\begin{figure}[!hbt]
  \centering

\begin{tikzpicture}
\node[anchor=south west,inner sep=0] (image) at (0,0) {

    \begin{subfigure}{0.49\textwidth}
    \centering
    \includegraphics[width=\linewidth]{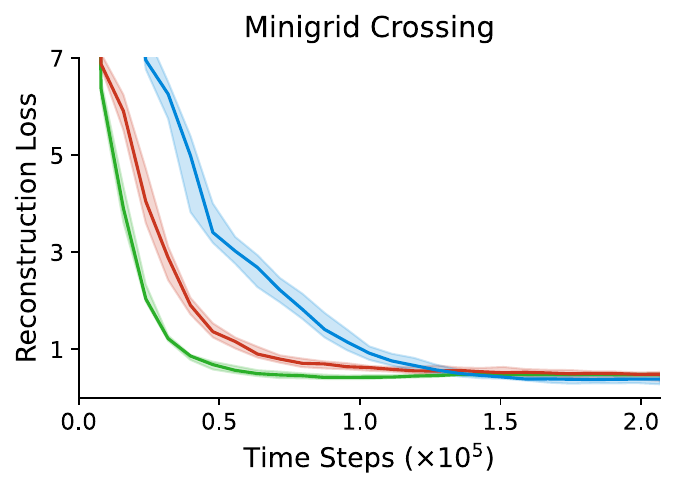}
  \end{subfigure}
  \begin{subfigure}{0.50\textwidth}
    \centering
    \adjustbox{trim={13} {0} {0} {0},clip}%
    {\includegraphics[width=\linewidth]{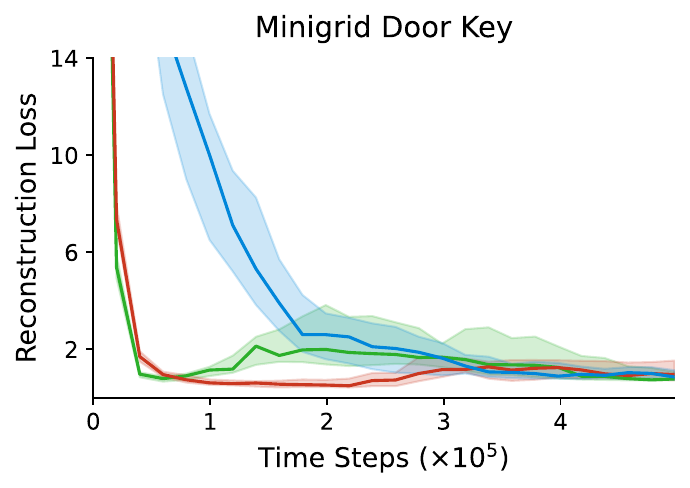}}
  \end{subfigure}

};
\begin{scope}[x={(image.south east)},y={(image.north west)}]
  \node [anchor=east, color=plotred, font=\plotfont] (cont) at (0.48, 0.28) {Vanilla AE};
  \node [anchor=east, color=plotblue, font=\plotfont] (disc) at (0.29, 0.5) {VQ-VAE};
  \node [anchor=east, color=plotgreen, font=\plotfont] (fta) at (0.155, 0.22) {FTA AE};
\end{scope}
\end{tikzpicture}
  
  \caption{Median reconstruction loss of the autoencoder during episodic RL training. The autoencoder is trained on observations randomly sampled from a buffer that grows as the RL training progresses. Lower is better, indicating a better reconstruction of the input observation. The plot depicts a 95\% confidence interval around the median over 30 runs. We plot the median of this metric as there are a few outliers that drastically skew the average. The VQ-VAE in particular exhibits the highest variance in reconstruction loss, but this does not seem to hinder the representation's performance in the RL setting.}
  \label{fig:vanilla_rl_recon}
\end{figure}

\begin{figure}[!h]
  \centering
  \begin{tikzpicture}
    \node[anchor=south west,inner sep=0] (image1) at (0,0) {
      \begin{subfigure}{0.265\textwidth}
        \centering
        \includegraphics[width=\linewidth]{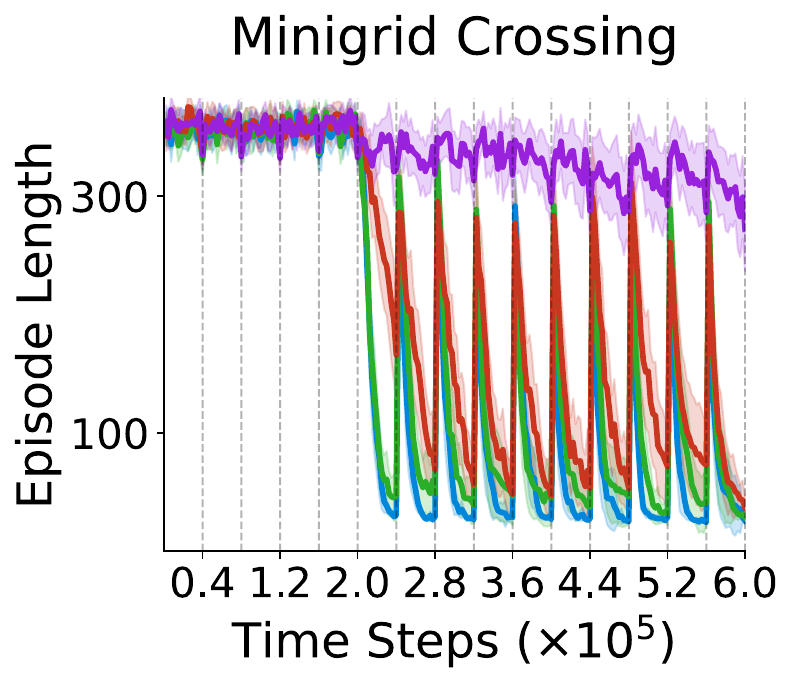}
        \caption{}
      \end{subfigure}
    };
    
    \begin{scope}[x={(image1.south east)},y={(image1.north west)}]
      \node [anchor=east, color=plotpurple, font=\plotfontvsmall] (rl) at (0.97, 0.865) {End-to-End};
    \end{scope}
    
    \hspace{-2.8mm}  
    \node[anchor=south west,inner sep=0] (image2) at (image1.south east) {
      \begin{subfigure}{0.275\textwidth}
        \centering
        \adjustbox{trim={10.5} {0} {0} {0},clip}%
        {\includegraphics[width=\linewidth]{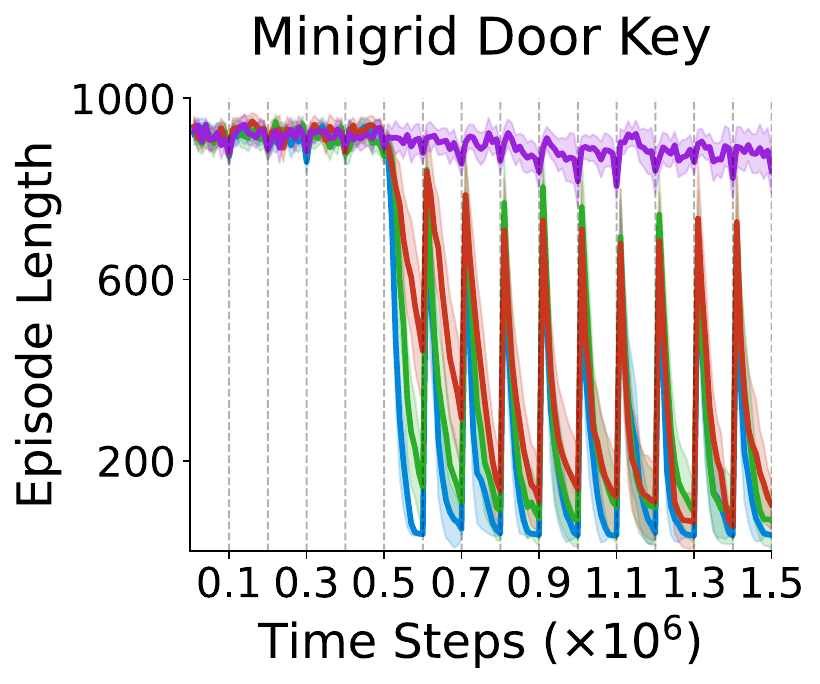}}
        \caption{}
      \end{subfigure}
    };
    
    \hspace{-2mm}  
    \node[anchor=south west,inner sep=0] (image3) at (image2.south east) {
      \begin{subfigure}{0.259\textwidth}
        \centering
        \includegraphics[width=\linewidth]{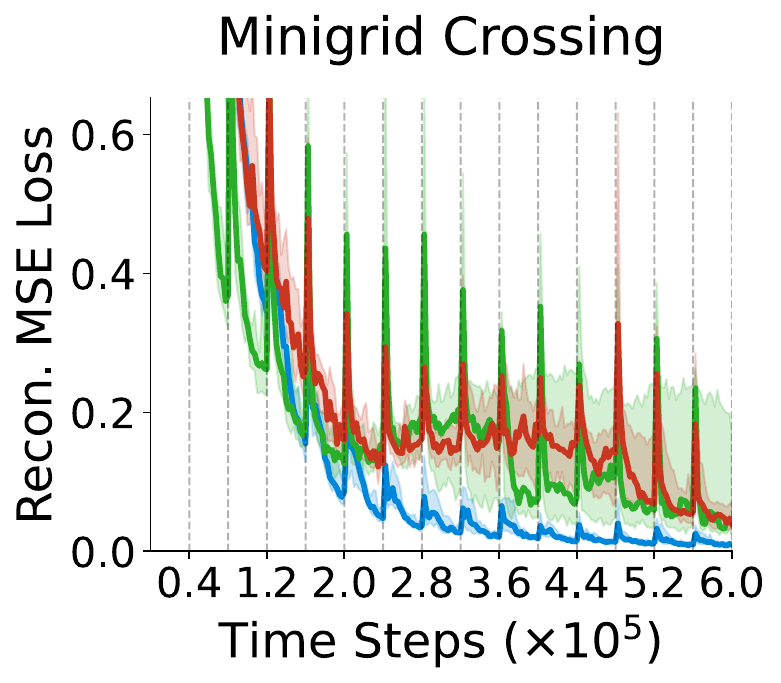}
        \caption{}
      \end{subfigure}
    };
    
    \begin{scope}[x={(image3.south east)},y={(image3.north west)}]
      \node [anchor=east, color=plotblue, font=\plotfontvsmall] (disc) at (0.625, 0.31) {VQ-VAE};
      \node [anchor=east, color=plotred, font=\plotfontvsmall] (cont) at (0.412, 0.8) {Vanilla AE};
      
      \draw [color=plotblue, line width=1pt] (disc.east) -- ++(0.005, 0.01); 
      \draw [color=plotred, line width=1pt] (cont.west) -- ++(-0.05, 0.01); 
    \end{scope}
    
    \hspace{-2mm}  
    \node[anchor=south west,inner sep=0] (image4) at (image3.south east) {
      \begin{subfigure}{0.26\textwidth}
        \centering
        \adjustbox{trim={9.5} {0} {0} {0},clip}%
        {\includegraphics[width=\linewidth]{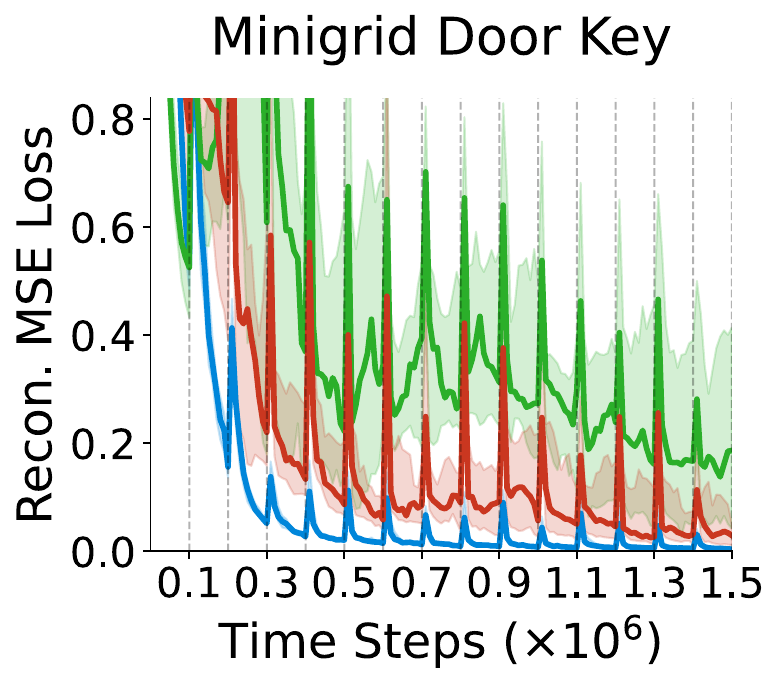}}
        \caption{}
      \end{subfigure}
    };
    
    \begin{scope}[x={(image1.south east)},y={(image4.north west)}]
      \node [anchor=east, color=plotgreen, font=\plotfontvsmall] (fta) at (1.78, 0.7) {FTA AE};
    \end{scope}
    
  \end{tikzpicture}

  \caption{(a-b) Mean agent performance as the environments change at intervals indicated by the dotted, gray lines. Lower is better. (c-d) Median encoder reconstruction loss. Lower peaks mean the representation generalizes better, and a quicker decrease means the autoencoder is learning faster. Overall, a lower reconstruction loss is better. (a-d) Results are averaged over 30 runs and depict 95\% confidence intervals. Performance is plotted after an initial delay to learn representations, after which all methods are trained with PPO.}
  \label{fig:full_changing_results}
\end{figure}
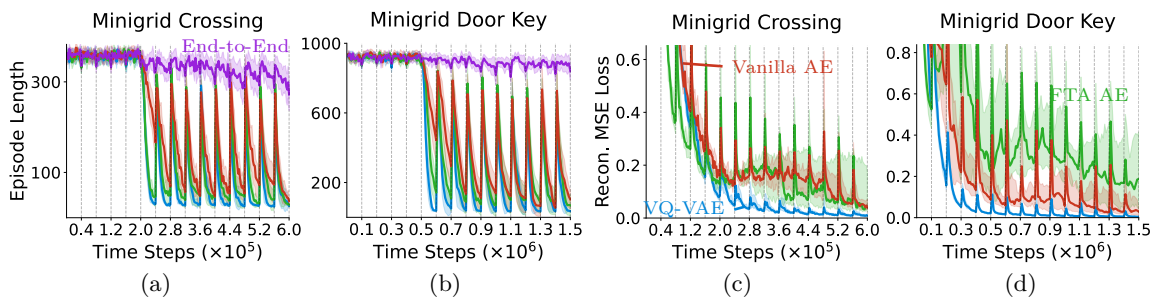

This section contains additional materials that help describe the RL training process and results. Algorithm \ref{alg:rl_training} provides pseudo-code for episodic and continual RL training. Figure \ref{fig:changing_envs} shows different environment variations used in the continual learning setting. Figure \ref{fig:vanilla_rl_recon} plots the reconstruction loss of the autoencoder during episodic RL training. And lastly, Figure \ref{fig:full_changing_results} depicts the full results of the continual RL runs starting from the first timestep.

\begin{algorithm}
\caption{Reinforcement Learning Training Process} \label{alg:rl_training}
\begin{algorithmic}

\STATE Initialize the encoder, $f_{\btheta}$, and decoder, $g_{\bphi}$
\STATE Initialize the policy and value networks, $\pi_{\bpsi}$ and $V_{\bpsi}$, with combined parameters $\bpsi$
\STATE $\mathcal{D} \leftarrow \emptyset$ \COMMENT{Dataset of observations}
\STATE Set number of interaction steps, $N$, batch size, $B_0$, autoencoder epochs, $L$, PPO epochs K, PPO start step $P$, and autoencoder batch size, $B_1$
\STATE For continual learning experiments, specify environment change frequency, $C$
\\
\WHILE{number of interactions is less than $N$}
    \STATE Enact policy $\pi_{\bpsi}$ in the environment to obtain a batch of $B_0$ transition tuples
    \IF{interaction step $\ge P$}
        \STATE Using the online data, perform $K$ epochs of PPO updates on parameters $\bpsi$
    \ENDIF
    \FOR{L steps}
        \STATE Sample a batch of observations $(s_0, s_1, ..., s_{B_1}) \in \mathcal{D}$
        \STATE Apply the autoencoder and calculate the reconstruction loss
        \STATE Update parameters $\btheta$ and $\bphi$ using Adam
    \ENDFOR

    \IF{doing continual learning and $C$ interaction steps have passed}
        \STATE Randomize the environment
    \ENDIF
    
\ENDWHILE

\end{algorithmic}
\end{algorithm}

\captionsetup[subfigure]{skip=1pt}

\end{document}